\documentclass[10pt,table]{article} 
\usepackage[accepted]{tmlr}

\usepackage{amsmath,amsfonts,bm}

\def\eqref#1{equation~\ref{#1}}

\def\1{\bm{1}}

\def\vc{{\bm{c}}}

\def\vh{{\bm{h}}}

\def\vn{{\bm{n}}}

\def\vq{{\bm{q}}}

\def\vx{{\bm{x}}}

\def\mA{{\bm{A}}}
\def\mB{{\bm{B}}}

\DeclareMathAlphabet{\mathsfit}{\encodingdefault}{\sfdefault}{m}{sl}
\SetMathAlphabet{\mathsfit}{bold}{\encodingdefault}{\sfdefault}{bx}{n}

\usepackage{hyperref}
\usepackage{url}

\usepackage{booktabs}
\usepackage{dblfloatfix}
\usepackage{xcolor}  
\usepackage{float}
\usepackage{wrapfig}
\usepackage{subcaption}  
\usepackage{enumitem}
\usepackage[capitalize,noabbrev]{cleveref}
\usepackage[utf8]{inputenc}
\usepackage{listings}

\usepackage[textsize=tiny]{todonotes}
\definecolor{LightGray}{gray}{0.9} 

\usepackage{ifthen}

\newboolean{showrevisions}
\setboolean{showrevisions}{false}

\newcommand{\revised}[1]{%
  \ifthenelse{\boolean{showrevisions}}%
    {\textcolor{blue}{#1}}%
    {#1}%
}

\title{\textcolor{blue}{A}\textcolor{red}{B}\textcolor[rgb]{0,0.5,0}{C}: \textcolor{blue}{A}chieving \textcolor{red}{B}etter \textcolor[rgb]{0,0.5,0}{C}ontrol of Visual Embeddings using VLLMs}

\author{\name Benjamin Schneider, \addr University of Waterloo \email Benjamin.Schneider@uwaterloo.ca
      \AND
      \name Florian Kerschbaum, \addr University of Waterloo \email florian.kerschbaum@uwaterloo.ca
      \AND
      \name Wenhu Chen, \addr University of Waterloo \email wenhuchen@uwaterloo.ca
      }

\def\month{08}  
\def\year{2025} 
\def\openreview{\url{https://openreview.net/forum?id=RezANmBpxW}} 

\begin{document}

\maketitle

\begin{abstract}
Visual embedding models excel at zero-shot tasks like visual retrieval and classification.
However, these models cannot be used for tasks that contain ambiguity or require user instruction.
These tasks necessitate an embedding model which outputs can use a natural language instruction to control the representation of a visual embedding. 
Existing CLIP-based approaches embed images and text independently, and fuse the result.
We find that this results in weak interactions between modalities, and poor user control over the representation.
We introduce \textbf{ABC}, an open-source multimodal embedding model that uses a vision-language model backbone to deeply integrate image features with natural language instructions.
\textbf{ABC} achieves best-for-size performance on MSCOCO image-to-text retrieval and is the top performing model on classification and VQA tasks in the Massive Multimodal Embedding Benchmark.
With a strongly unified vision-language representation, \textbf{ABC} can use natural language to solve subtle and potentially ambiguous visual retrieval problems.
To evaluate this capability, we design \texttt{CtrlBench}, a benchmark that requires interleaving textual instructions with image content for correct retrieval.
\textbf{ABC} advances the state of visual embeddings, outputting high-quality visual representations with natural language control.
Our model and datasets are available at our project page:
\end{abstract}

\vspace{-0.5cm}
\begin{center}
    \href{https://tiger-ai-lab.github.io/ABC/}{\texttt{https://tiger-ai-lab.github.io/ABC/}}
\end{center}
\vspace{-0.5cm}

\section{Introduction}

\begin{wrapfigure}{r}{0.35\textwidth}
\vspace{-13pt}
  \centering
\includegraphics[width=0.35\textwidth]{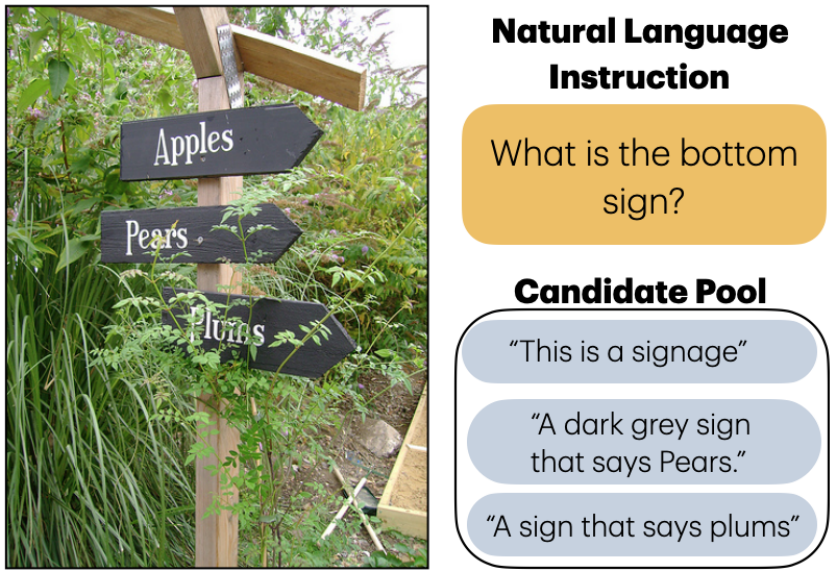}
\caption{{A task from \texttt{CtrlBench}. Conditioning on the natural language instruction is required to unambiguously retrieve a candidate.}}  \label{fig:intro_top}
\vspace{-10pt}
\end{wrapfigure}

Visual embeddings have become a foundational representation in computer vision.
Image embedding models have become the state of the art for many zero-shot tasks, including visual retrieval~\citep{internvl_1} and image classification~\citep{coca}.
Since the release of CLIP~\citep{clip}, its dual encoder architecture has remained the state of the art for producing high-quality visual embeddings~\citep{coca, eva-clip, internvl_1}.
However, CLIP only supports \emph{separately} embedding images {or} text~\citep{clip}.
Therefore, complex visual embedding tasks which require additional specification are impossible.
For example, a CLIP model cannot distinguish which is the best answer in \cref{fig:intro_top}.
\revised{
All candidates are plausible unless the user provides additional instruction.
However, additional conditioning on a natural language question \emph{disambiguates} the query, making it apparent that the bottom candidate is the best answer~\citep{stelmakh-etal-2022-asqa}.
Conditioning on natural language gives the user greater control over their visual query.
This is especially critical for retrieval augmented generation systems, which must unambiguously retrieve a small number of highly relevant facts from large knowledge bases, to support accurate text generation~\citep{distract}.
}

We find that existing approaches suffer from two problems:
\textbf{(1)} Reliance on generic or human-annotated instructions~\citep{vlm2vec,gme}.
Unlike the abundant noisy image-text data used to train CLIP models~\cite{laion,c12m,conceptualcaptions}, datasets containing multiple modalities and diverse instruction phrases are significantly more scarce and challenging to curate at scale.
This is problematic for embedding models, which require massive datasets for the best representations~\citep{open_clip}.
Often vague instructions are used repetitively during training, resulting in overfitting of the instruction~\citep{uniir, ins_overfit1}.
\textbf{(2)} Weak interaction between modalities.
Previous works fuse embeddings outputted by CLIP models~\cite{magiclens, uniir}.
This approach prevents deeper interaction between modalities, resulting in superficial use of the instructions~\cite{vlm2vec}.

\begin{figure*}
    \centering
    \includegraphics[width=1.\linewidth]{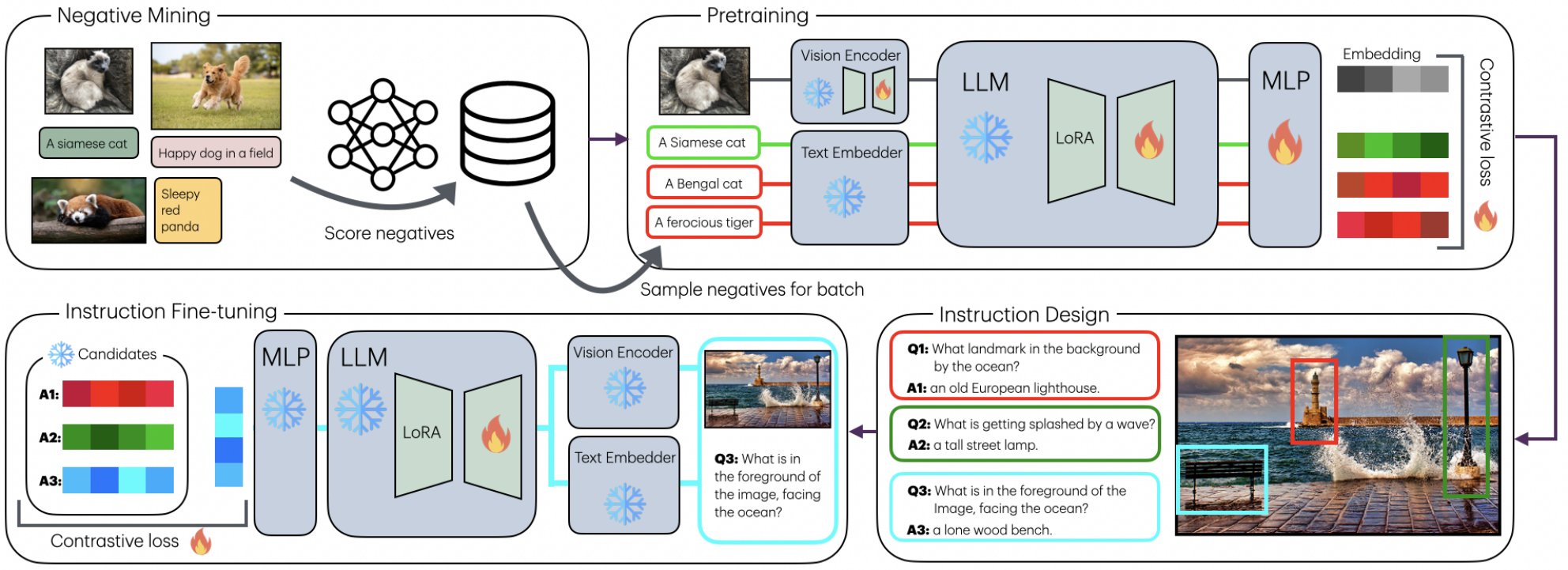}
    \caption{An overview of \textbf{ABC}'s training regime. First, we mine negatives to augment each batch in our pretraining dataset with \emph{almost plausible} text candidates. Then we pretrain our model using contrastive loss on independently embedded text-image pairs. Once our model has learned to output high-quality uncontrolled visual embeddings, we introduce instructions. We create multiple instructions and candidate captions for a single image, with each one focusing on different aspects of the image. Lastly, we fine-tune using multiple  captions for same image, the positive caption for each query corresponds to the natural language instruction. Therefore, \textbf{ABC} uses the instruction to distinguish the correct candidate for the query.}
    \label{fig:intro}
    \vspace{-20pt}
\end{figure*}

To this end, we introduce \textbf{ABC}, a model that uses natural language instructions to control visual embeddings.
\textbf{ABC}'s Vision Large language Model (VLLM) backbone allows it to integrate natural language instructions when crafting visual embeddings.
We find that training our model has two fundamental challenges:
\textbf{(1)} Extracting useful contrastive embeddings from a pretrained generative VLLM.
\textbf{(2)} Designing an instruction fine-tuning method that lets users modify multimodal embeddings using natural language instructions.
To train \textbf{ABC}, we adopt a multi-stage training process.
In the initial pretraining stage, we use contrastive training with carefully selected negatives to develop a model that generates embeddings, similar to CLIP.
In the instruction fine-tuning stage, we train a lightweight adapter to project queries consisting of images and natural language instructions into the embedding space of the pretrained model.
Our synthetic instructions to correspond to different aspects \emph{of the same image}, therefore, \textbf{ABC} learns to use the instruction for fine-grained control.
Our training results in a model that produces powerful and controllable visual embeddings.

\textbf{ABC} achieves impressive zero-shot performance in retrieval, classification, and visual question answering (VQA) tasks.
In MSCOCO~\citep{mscoco} image-to-text retrieval, our model outperforms all CLIP models containing at most $8$ billion parameters.
Furthermore, our model outperforms all other models on the zero-shot classification and VQA splits of MMEB~\citep{vlm2vec}, a multimodal embedding benchmark spanning 19 tasks.
Lastly, we design \texttt{CtrlBench} to measure our model's ability to use natural language instructions to control retrieval.
Using \texttt{CtrlBench}, we show that \textbf{ABC} can accomplish visual retrieval tasks that are fundamentally ambiguous without utilizing natural language instructions.

Our contribution is threefold.
\textbf{(1)} \textbf{ABC}: an open-source multimodal embedding model that uses natural language instructions to control visual embeddings.
We demonstrate that \textbf{ABC} produces powerful embeddings by benchmarking on zero-shot tasks \emph{and} robustly utilizes natural language instructions to control embeddings.
We study how key design choices—including vision encoder resolution, contrastive loss temperature, and VLLM backbone—impact embedding quality.
\textbf{(2)} We provide a \emph{decoupled} methodology for adapting VLLMs into dense embedding models.
Previous work integrates instructions during the large-scale contrastive training run.
We show that these stages can be decoupled, resulting in a lightweight and adaptable fine-tuning stage that requires only $100$ training steps using synthetic instructions.
\textbf{(3)} Lastly, we introduce \texttt{CtrlBench}, a novel benchmark for measuring instruction-controlled retrieval.
\texttt{CtrlBench} \emph{requires} the model to interleave modalities to retrieve the correct response.

\section{Background}

\revised{\textbf{Visual embeddings.}
\citet{clip} proposed CLIP, which projects images and text to a shared dense embedding space.
Due to its contrastive training, the similarity of vectors is computed as simple dot product~\citep{contrastive_loss}.
CLIP embeddings are used for tasks such as zero-shot image classification and image-to-text retrieval, by computing the similarity of an image query $\vq$ with a set of candidates $\vc \in \mathcal{C}$.
An embedding model $f$ maps queries and candidates to vectors in $\mathbb{R}^n$.
Given a query $\vq$ and a set of candidates $\mathcal{C} \subseteq \mathbb{R}^n$ containing the correct answer $\vc^+$, the retrieved candidate is:
\[
\vc^* = \arg\max_{\vc \in \mathcal{C}} \; f(\vq) \cdot f(\vc)
\]
}
\revised{
The model is succeeds if $\vc^* = \vc^+$.
Alternatively, when measuring metrics such as recall@K, K of the top scoring candidates are returned.
Subsequent works have identified several key factors for training training high-quality CLIP models:
DataComp~\citep{datacomp} demonstrated the importance of data choice during to create high-quality embeddings.
\citep{scaling_laws} showed that large-scale pretraining and batch size are crucial for producing the best performing models.
Lastly, Align~\citep{align} found that optimizing the setting of the temperature $\tau$, used to scale the contrastive loss function, was crucial.
}

\revised{\textbf{Multimodal embeddings for Visual Question Answering.} 
Visual embeddings, especially those of diverse scenes containing many objects, offer limited control over the embedding~\citep{uniir}.
However, a {multimodal embedding} $\vq = (q_i, q_t)$, jointly conditioned on image $q_i$ and text instruction $q_t$ allows for flexible user natural language control over embeddings.
This allows the user to create embeddings corresponding to specific objects or concepts in an image~\citep{vlm2vec}.
For a VQA task, the model embeds $(q_i, q_t)$ jointly into a dense vector.
The model uses cosine similarity to retrieve the maximally similar text candidate $c^* \in \mathcal{C}$.
As in the unconditioned setting, the model is successful if $c^*$ the correct answer. 
We use the VQA split of MMEB~\citep{vlm2vec}, which contains 10 tasks that span text-conditioned classification and retrieval.
We also introduce \texttt{CtrlBench}, a benchmark designed to measure text-conditioned visual retrieval when image-only retrieval is \emph{ambiguous}.}

\revised{\textbf{Ambiguity in visual retrieval.}
Ambiguity is a well studied problem in information retrieval and natural language processing~\citep{10.1145/3534965}. Often, users provide queries that have multiple distinct, nevertheless plausible answers.
Several rectifying approaches have been proposed, including using follow-up questions to disambiguate queries~\citep{kim2023treeclarificationsansweringambiguous, Aliannejadi_2019} and query refinement~\citep{stelmakh-etal-2022-asqa}.
Following \citet{min2020ambigqaansweringambiguousopendomain}, we define an image query $q_i$ as ambiguous if its corresponding candidate pool $\mathcal{C}$ contains multiple semantically distinct plausible answers.
Additional conditioning, such as on a natural language instruction $q_t$, disambiguates $q_i$ if the joint query $\vq = (q_i, q_t)$ has exactly one plausible answer in $\mathcal{C}$~\citep{stelmakh-etal-2022-asqa}.
}

\revised{\textbf{Extracting Embeddings from LLMs.}
Generative large language models are a source of pretrained bases training dense embeddings models.
NV-Embed~\citep{nvembed} used contrastive training on an LLM backbone to achieve the best performance on the MMEB text embedding benchmark~\citep{mteb}.
Several architectural modifications have been proposed for adapting LLMs to produce dense embeddings~\cite{improvedtextembeddings,llm2vec,nvembed}.
LLM2VEC~\cite{llm2vec} changed the attention mask to be bidirectional to allow information flow between all tokens.
Furthermore, several methods for extracting dense embedding from LLMs hiddens have been proposed.
These include picking the last token~\citep{vlm2vec}, mean token pooling~\citep{llm2vec} or an additional attention layer~\citep{nvembed}.
}

\section{ABC}

\Cref{fig:intro} is an overview of our training regime and model architecture.
Our training regime consists of $2$ distinct stages; pretraining and instruction fine-tuning.
The pretraining stage trains on image-caption pairs to adapt features used for generative modeling into features for dense embeddings.
As our pretraining does not require instructions, it can be easily scaled using any large image-captioning data source~\citep{c12m, laion}.
\revised{In the second stage, we train an adapter to allow \textbf{ABC} to embed images conditioned on a text instruction, enabling natural language control over visual representations. Using contrastive learning with synthetic instructions, the \textbf{ABC} learns to represent different visual aspects of the same image based solely on textual commands.}

\subsection{Model Design}

\textbf{Pretraining Architecture}. During pretraining images and text are embedded independently, and aligned using contrastive loss, much like CLIP~\citep{clip}.
Text and images are converted to hiddens $\vh^{(0)} \in \mathbb{R}^{s \times d}$ using the embedding layer and vision encoder, respectively.
The LLM module is shared between both modalities.
To adapt the LLM to output dense embeddings, we make several architectural changes.
Following \citet{llm2vec}, we enable bidirectional attention, allowing all tokens to attend to all other tokens.
To create our dense embeddings, we average over the sequence dimension in the last hidden layer $\vh^{(l)}$ (\eqref{eq:mlp1}), and project the result $\vx \in \mathbb{R}^{d}$ using a residually connected $MLP$ layer (\eqref{eq:mlp2}).

\begin{align}
    \vx         &= \frac{1}{s} \sum_{i}^{s} \vh^{(l)}_i  \label{eq:mlp1} \\
    MLP(\vx)    &= \vx + \mA g( \mB \vx ) \label{eq:mlp2}
\end{align}

Where $\mA$ and $\mB$ are parameter matrices and $g$ is the element-wise SELU function~\citep{selu}.
To train our backbone, we apply LoRA~\citep{lora} adapters on both the vision encoder and LLM modules.
We optimize the contrastive loss temperature hyperparameter $\tau$ during pretraining.

\textbf{Instruction Fine-tuning Architecture}.
After pretraining, \textbf{ABC} can project queries consisting of images \emph{or} text into a shared embedding space. 
However, to control the representation of visual embeddings, we must project queries consisting of images \emph{and} text instructions into the embedding space.
To accomplish this, we train a low-rank LoRA adapter on the LLM component.
\revised{The LoRA hooks of the instruction adaptor are attached to the MLP and attention layers in each decoder block of the LLM backbone, we freeze all other weights during instruction fine-tuning.}
The adapter contrastively aligns image and instruction queries with text candidates embedded using the pretrained model.
For queries containing both an image and text instruction, we adopt the instruction template from our VLLM backbone~\citep{qwenvl}:

\begin{align}
    \texttt{<vision\_start>} \ \texttt{<image\_tokens>} \ \texttt{<vision\_end>} \ \texttt{<im\_start>} \ \texttt{Instruction:\;\{text\}} \ \texttt{<im\_end>}
\end{align}

We replace $\texttt{<image\_tokens>}$ with tokens output by vision encoder.
$\texttt{<vision\_start>}$ and $\texttt{<vision\_end>}$ are special vocabulary tokens used to denote the start and end of the sequence of image tokens.
Similarly, $\texttt{<im\_start>}$ and $\texttt{<im\_end>}$ are used to denote the start and end of text tokens.
If no natural language instruction is used, the last token is $\texttt{<vision\_end>}$.
We do not back-propagate through text candidate embeddings, only the queries containing an image and instruction.
This ensures that our image-text queries share the same features as text and images embedded using the pretrained model.
We can easily alternate between embedding with or without instructions by disabling the instruction fine-tuned LoRA adapter.
We freeze the temperature $\tau$ during fine-tuning.

\subsection{Pretraining Data}
\label{sec:data} 

\begin{wrapfigure}{r}{0.35\textwidth}
  \centering
\includegraphics[width=0.35\textwidth]{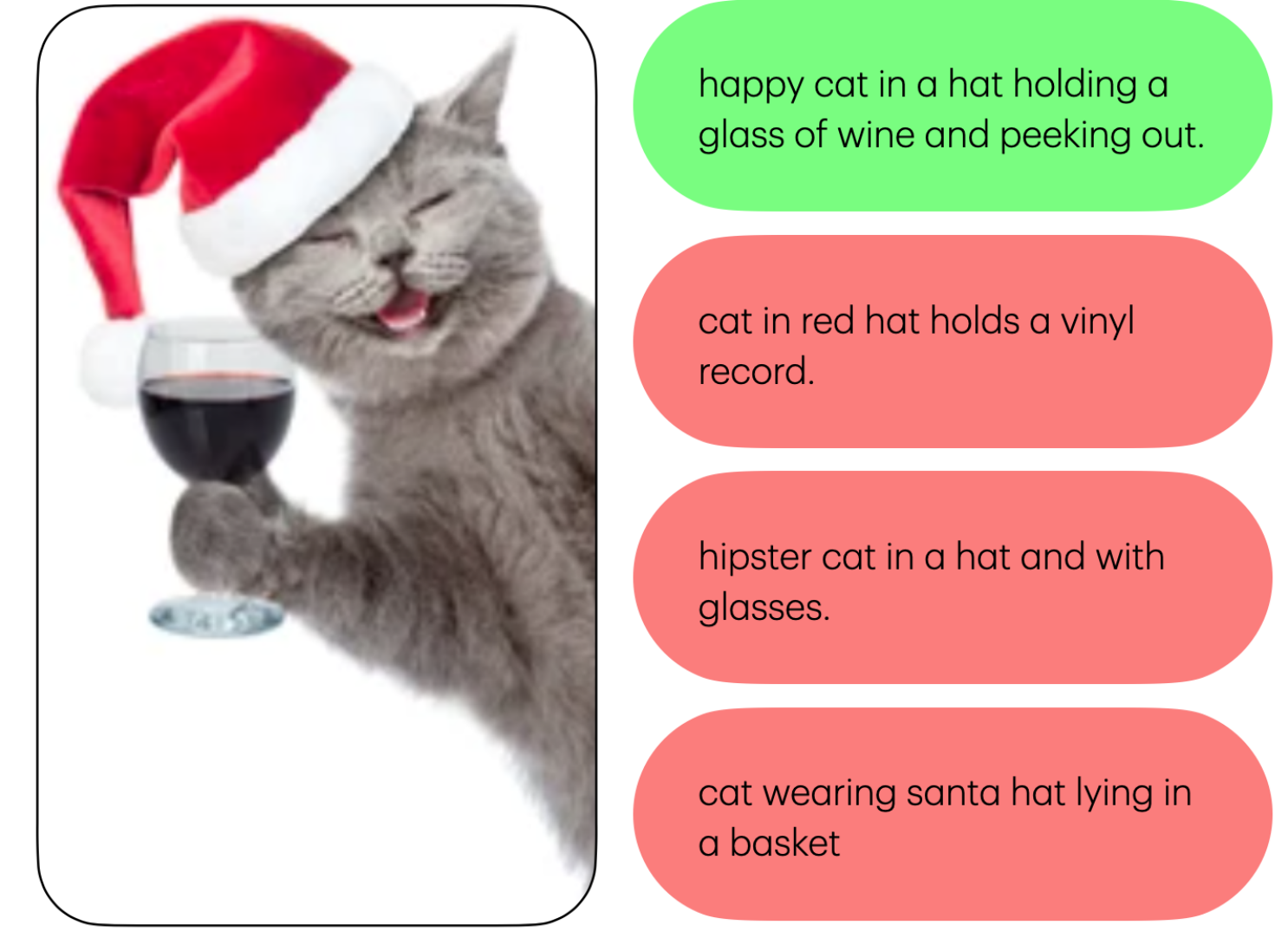}
\caption{The positive caption \textcolor{green}{(green)} is the best caption for the image. The mined negatives \textcolor{red}{(red)} are relevant but not the best choice.}  \label{fig:sample}
\vspace{-10pt}
\end{wrapfigure}

To create our pretraining dataset we employ negative mining on Conceptual Captions~\citep{conceptualcaptions}.
We derive the mined negatives for our dataset as follows:
\textbf{(1)} We do a small pretraining run using only in-batch negatives.
\textbf{(2)} We use the resulting model to calculate similarity scores between all images and captions in our pretraining dataset.
This approach avoids a circular dependence on a third-party embedding model to train our embedding model.
Therefore, it is easily extensible to modalities where an existing embedding model is not publicly available.
\textbf{(3)} To prevent our negatives from being too similar to our positive samples, we set a similarity threshold $\epsilon \in [0, 1]$.
We only sample negatives that have a similarity score of at most $\epsilon$ times the similarity score of the correct candidate.
We randomly choose our mined hard negatives from the $100$ candidate captions below the threshold.
%
%
This results in the text candidates shown in \cref{fig:sample}.
The mined text negatives are clearly relevant, but the correct caption is still the best answer.

In our pretraining run, each batch consists of $N$ image queries (without instructions) and $M$ text candidates.
We include $M-N$ mined text negatives in each batch.
Therefore, each image query has $\frac{M}{N} - 1$ corresponding mined negatives.
Our pretraining loss function is given by \Cref{eq_loss}.

\begin{equation}\label{eq_loss}
\mathcal{L}_{pretrain} = -\sum_{i}^{N}\log \frac{\exp(f(\vq_{i}) \cdot f(\vc_i^+) / \tau)}{\sum_{j=1}^{N}(\exp(\frac{f(\vq_{i}) \cdot f(\vc_j^+)}{\tau}) + \sum_{k=1}^{\frac{M}{N}-1}\exp(\frac{f(\vq_{i}) \cdot f(\vn_j^k)}{\tau}))}
\end{equation}

For a given query $\vq_i$, $\vn_i^k$ is the $k_{th}$ mined hard negative candidate and $\vc_i^+$ is its positive candidate.
$\tau$ is the temperature hyperparameter used to scale the loss.
$\vn_j^k$ is a mined negative for $\vx_i$ when $i=j$, otherwise it acts as a regular in-batch negative.
The embeddings are unit normalized before loss is computed.
We scale the loss by $\frac{1}{N}$, the number of image queries in the batch.

\subsection{Instruction Fine-tuning Data}

To create our instruction fine-tuning dataset we use Visual Genome~\citep{visualgenome}, a dataset comprised of images with captioned bounding boxes.
When choosing which bounding boxes to use for each image, we filter by the size of the bounding box (width $\times$ height).
We exclude the $5$ largest bounding boxes, as we find they often do not represent specific objects or aspects of the scene.
For each image, we randomly choose $4$ bounding boxes and their respective captions.
We then prompt GPT-4o~\citep{gpt4o} to create instructions corresponding to each bounding box caption.
We sample multiple captions from each image so that they can be used as negatives for each other during instruction fine-tuning.
This ensures that each image query has multiple realistic captions in its batch.
Therefore, \emph{utilizing the instruction is required to choose the correct text candidate unambiguously.}
We provide the prompt and generation settings used to create our instruction fine-tuning dataset in \Cref{apx:prompt}.
We instruction fine-tune using exclusively in-batch negatives.
However, we group all queries that contain the same image into the same batch.
This ensures that queries always have multiple relevant text candidates. 

\section{Experiments}

\begin{table*}[!b]
    \centering

\resizebox{1\textwidth}{!}{%
\begin{tabular}{l|cccccc|cccccc}
    
    \toprule 
    & \multicolumn{6}{c|}{\textbf{MSCOCO} (5K test set)} & \multicolumn{6}{c}{\textbf{Flickr30K} (1K test set)} \\
    & \multicolumn{3}{c}{Image $\rightarrow$ Text} & \multicolumn{3}{c|}{Text $\rightarrow$ Image} & \multicolumn{3}{c}{Image $\rightarrow$ Text} & \multicolumn{3}{c}{Text $\rightarrow$ Image} \\
    \cmidrule(lr){2-4} \cmidrule(lr){5-7} \cmidrule(lr){8-10} \cmidrule(lr){11-13}
    Model & R@1 & R@5 & R@10 & R@1 & R@5 & R@10 & R@1 & R@5 & R@10 & R@1 & R@5 & R@10 \\
    \midrule
    CLIP~\citep{clip}  & 58.4 & 81.5 & 88.1 & 37.8 & 62.4 & 72.2 & 88.0 & 98.7 & 99.4 & 68.7 & 90.6 & 95.2 \\
    ALIGN~\cite{align} & 58.6 & 83.0 & 89.7 & 45.6 & 69.8 & 78.6 & 88.6 & 98.7 & 99.7 & 75.7 & 93.8 & 96.8 \\
    FLAVA~\citep{flava} & 42.7 & 76.8 & - & 38.4 & 67.5 & - & 67.7 & 94.0 & - & 65.2 & 89.4 & - \\
    FILIP~\textbf{*}~\citep{filip} & 61.3 & 84.3 & 90.4 & 45.9 & 70.6 & 79.3 & 89.8 & {99.2} & \underline{99.8} & 75.0 & 93.4 & 96.3 \\
    CoCa~\textbf{*}~\citep{coca} &  66.3  &  86.2  &  91.8  &  \underline{51.2} &  74.2 & 82.0 & 92.5 &  \bf99.5 &  \bf99.9 &  \bf80.4 &  \bf95.7 & \bf97.7 \\
    OpenCLIP-G~\citep{open_clip} &  67.3  &  86.9  &  92.6  & \bf51.4 & \underline{74.9} &  \bf83.0 &  \underline{92.9} &  99.3 &  \underline{99.8} & \underline{79.5} &  \underline{95.0} & \underline{97.1} \\
    EVA-02-CLIP-E+~\citep{eva-clip} &  \underline{68.8}  &  \underline{87.8}  &  \underline{92.8}  & 51.1 & \bf75.0 &  \underline{82.7} &  \bf93.9 &  \underline{99.4} &  \underline{99.8} & 78.8 &  94.2 & 96.8 \\
    \midrule
    \rowcolor{LightGray} ABC (ours) & \bf 69.2 & \bf 87.9 & \bf 93.2 & {47.6} & {72.1} & {80.6} & {90.7} & 99.0 & 99.5 & 74.6 & 92.6 & 95.45\\
    \bottomrule
  
\end{tabular}

}
\caption{Comparison of retrieval performance on MSCOCO~\cite{mscoco} and Flickr30K~\cite{flickr} datasets (Karpathy split). Best performance is \textbf{bold}, second best is \underline{underlined}. \textbf{*} indicates a closed-weight model.}
\label{tab:retrieval_results}
\end{table*}

We evaluate two aspects of the model.
As controlling a poor-quality embedding is useless, we first assess the quality of embeddings output by \textbf{ABC}.
We evaluate our model on a variety of zero-shot retrieval and classification tasks~(\cref{sec:rep-quality}).
We study how variable resolution capabilities~(\cref{sec:resolution}), the settings for $\tau$ when contrastively training a VLLM~(\cref{sec:pretrain-loss}) and VLLM backbone choice (\cref{sec:backbone}) effect embedding quality.
{Secondly, we measure \textbf{ABC}'s ability to use natural language to control visual embeddings on several visual question answering tasks~(\cref{sec:control}).
Motivated by shortcomings in VQA benchmarks, we construct \texttt{CtrlBench}. 
A benchmark designed to require the model to jointly use the image and instruction to retrieve the most appropriate caption.}

\subsection{Training Settings}
We initialize \textbf{ABC} using Qwen2-VL-7B ~\citep{qwenvl}.
We train our negative mining model using only in-batch negatives with a batch size of $256$ for $1000$ steps.
\revised{We set $\epsilon=0.95$ and sample 7 mined negatives for each image. Mined negatives are sampled across the pretraining dataset.}
We pretrain using batches of $512$ image queries and $4096$ text candidates sharded across 8 NVIDIA A100-SXM4-80GB GPUs~\citep{batchgather} for 4000 steps.
We use a LoRA adapter with a rank of $64$ and a fixed alpha of $128$.
We limit the number of tokens output by the vision encoder to $512$ during training.
For our optimizer, we use AdamW~\citep{adamw} with a learning rate of $4 \times 10^{-5}$, betas of $0.9$ and $0.999$ and a weight decay of $10^{-3}$.
We warm-up for 3\% of training steps and initialize the temperature $\tau$ as $7 \times 10^{-2}$.
In our instruction fine-tuning stage, we use a lower rank LoRA adapter.
\revised{We set the rank and alpha to $16$ and $32$, respectively.}
Our instruction fine-tuning stage can be short, as our VLLM backbone is already instruction fine-tuned.
Therefore, we only instruction fine-tune for $100$ steps.
Each batch contains 128 unique images, with each image appearing four times, paired with a different instruction and a corresponding positive text candidate.

\textbf{Modifications to save VRAM.}
Due to our use of LoRA~\cite{lora}, the VRAM requirements for gradients and optimizer state is relatively low.
Consequentially, the model activations used by autograd for the backward pass account for most of VRAM used during training.
To address this, we make aggressive use of activation checkpointing, recomputing the activations of each decoder block during the backward pass.
Furthermore, we modify the VLLM backbone to skip the logits and cross-entropy loss computation.
With the large vocabulary size of modern LLMs, the output layer has become the most memory-intensive layer~\citep{ce_memory}.
As we only use the hidden state for calculating embeddings, the logits tensor is unnecessary.
We find that this simple change saves up to 11 GB of memory per device, allowing us to use to a significantly larger batch size.

\subsection{Inference Settings}
\revised{
When evaluating ABC we adjust the vision encoder to output a number of tokens corresponding to the resolution of the image.
We rescale images to the nearest resolution that fits our vision encoder, which has a $14\times14$ pixel patch size~\citep{qwenvl}.
For very high resolution images, we cap the number of visual tokens used to encode an image to $1024$, down-scaling the image accordingly.
When evaluating models that supports variable resolution, we vary their token budgets similarly~\citep{vlm2vec,gme}.
However, when we attempt to increase token budgets in CLIP models using the same method, we find that it degrades performance (\Cref{apx:infernece}).
We find that these models perform best when evaluated using their training resolution.
Therefore, we do not scale the token budget when evaluating CLIP models.
This includes instruction-finetuned CLIP models such as Uniir and MagicLens~\citep{uniir,magiclens}.}

\begin{table*}[!t]
\centering
\resizebox{1\textwidth}{!}{
\begin{tabular}{lcccccccc}
\toprule
& \textbf{CLIP} & \textbf{OpenCLIP} & \textbf{SigLIP} & \textbf{UniIR} & \textbf{MagicLens} & \textbf{BLIP2} & \textbf{MMRet} & \textbf{ABC (ours)} \\
\midrule
\rowcolor{LightGray} \textbf{Classification (9 tasks)} & & & & & & & & \\
ImageNet-1K~\cite{imagenet}       
& 55.8 & \underline{63.5} & 45.4 & 58.3 & 48.0 & 10.3 & 49.1 & \textbf{71.2} \\
HatefulMemes~\cite{hateful-memes} 
& 51.1 & 51.7 & 47.2 & \textbf{56.4} & 49.0 & 49.6 & 51.0 & \underline{52.1} \\
VOC2007~\cite{voc-2007}           
& 50.7 & 52.4 & 64.3 & 66.2 & 51.6 & 52.1 & \underline{74.6} & \textbf{81.4} \\
SUN397~\cite{sun397}              
& 43.4 & \underline{68.8} & 39.6 & 63.2 & 57.0 & 34.5 & 60.1 & \textbf{71.8} \\
Place365~\cite{place365}          
& 28.5 & \underline{37.8} & 20.0 & 36.5 & 31.5 & 21.5 & 35.3 & \textbf{40.7} \\
ImageNet-A~\cite{imagenet-a}      
& 25.5 & 14.2 & \underline{42.6} & 9.8 & 8.0 & 3.2 & 31.6 & \textbf{49.4} \\
ImageNet-R~\cite{imagenet-r}      
& 75.6 & \underline{83.0} & 75.0 & 66.2 & 70.9 & 39.7 & 66.2 & \textbf{86.8} \\
ObjectNet~\cite{objectnet}        
& 43.4 & \underline{51.4} & 40.3 & 32.2 & 31.6 & 20.6 & 49.2 & \textbf{67.7} \\
Country-211~\cite{clip}           
& \textbf{19.2} & 16.8 & 14.2 & 11.3 & 6.2 & 2.5 & 9.3 & \underline{18.5} \\
\midrule
\textit{All Classification} 
& 43.7 & \underline{48.8} & 43.2 & 44.5 & 39.3 & 26.0 & 47.4 & \textbf{60.0} \\
\bottomrule
\end{tabular}
}
\caption{Zero-shot classification results on MMEB~\citep{vlm2vec}. We compare with pretrained CLIP models~\citep{clip, open_clip}, instruction finetuned models derived from CLIP~\citep{uniir, magiclens} and other LLM backbone approaches~\citep{blip2,megapairs}.}
\label{tab:classification_results}
\end{table*}

\begin{figure*}[!b]
    \centering
    \hfill
    \begin{minipage}[t]{0.43\textwidth}
        \centering
        \includegraphics[width=\textwidth]{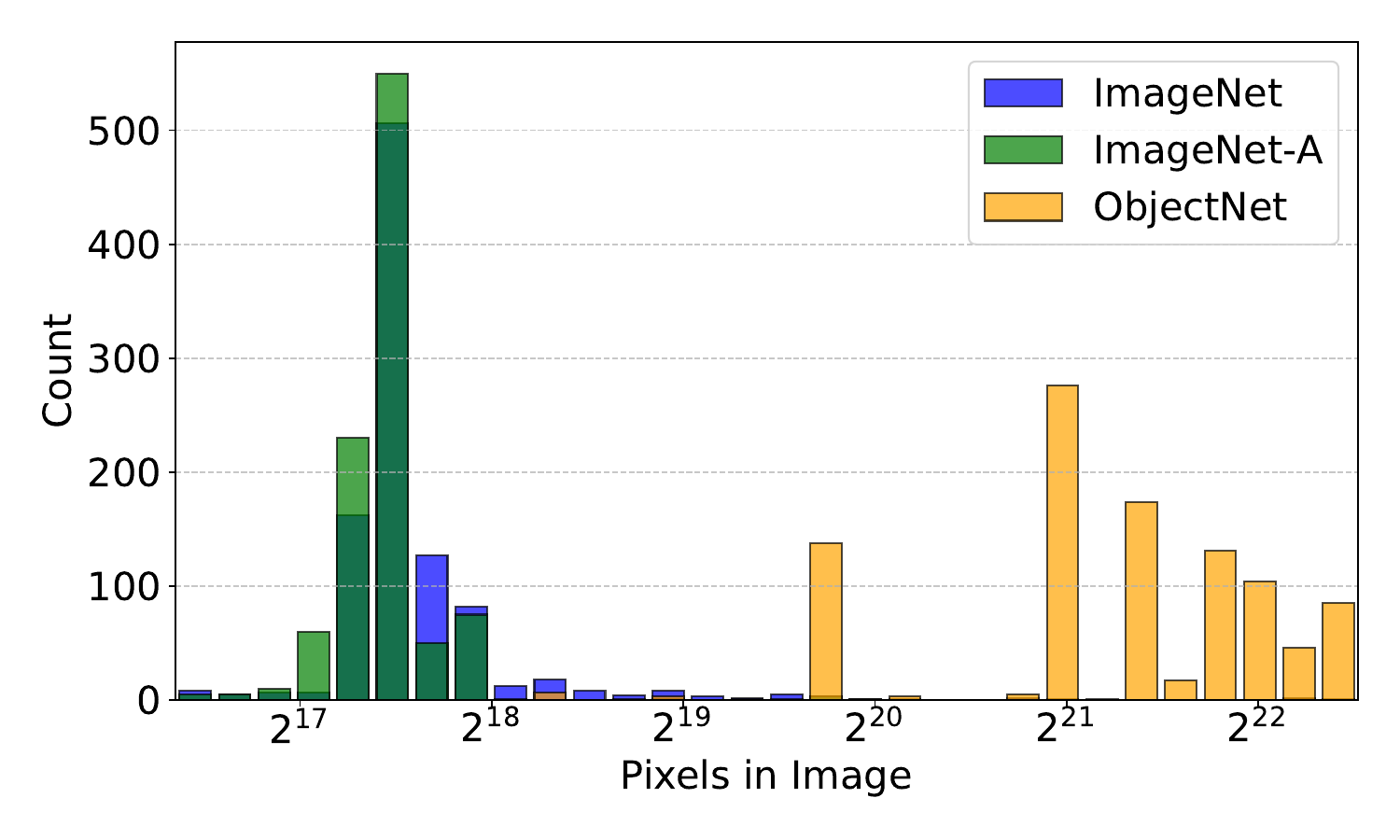}
    \end{minipage}
    \hfill
    \begin{minipage}[t]{0.43\textwidth}
        \centering
        \includegraphics[width=\textwidth, height=5cm]{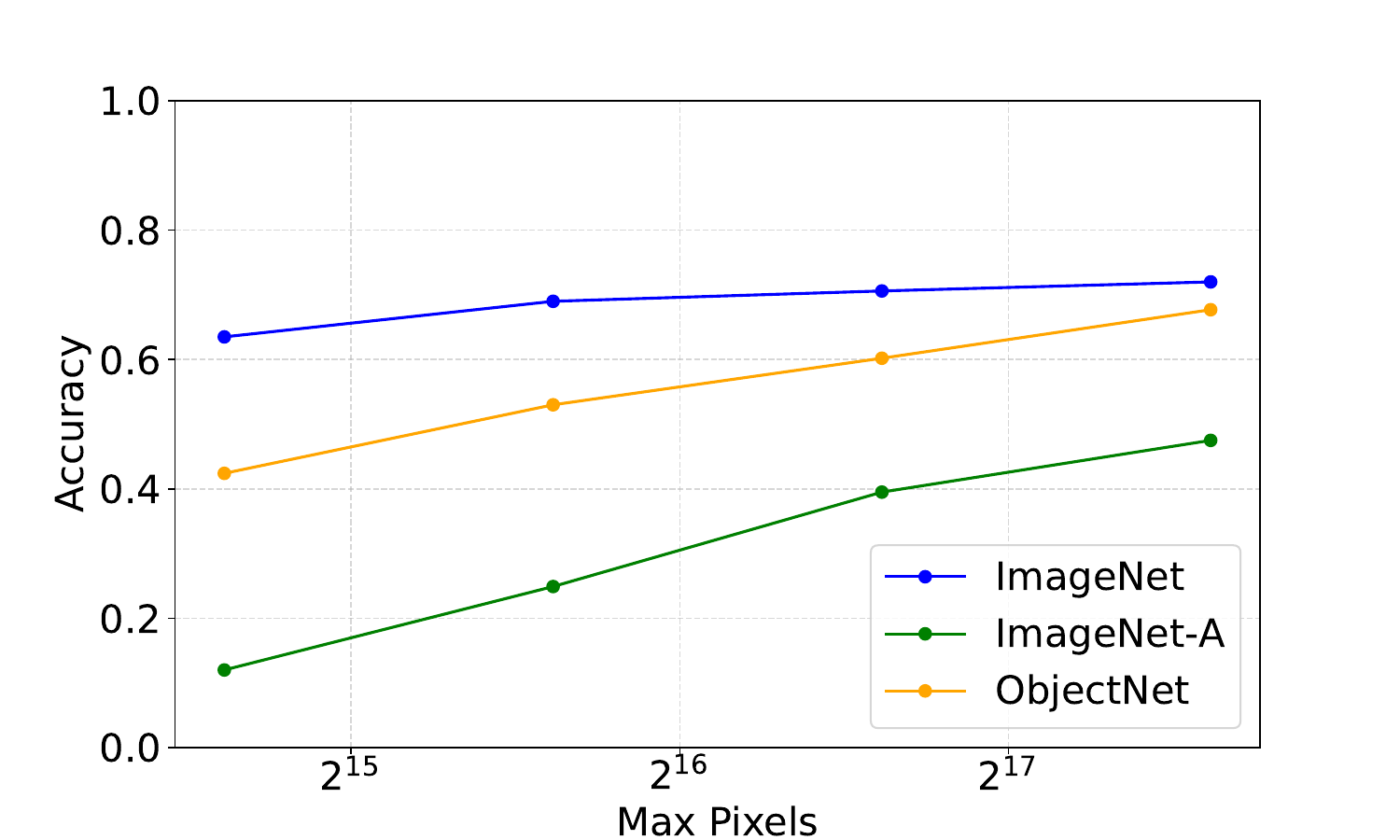}
    \end{minipage}
    \caption{\textbf{(Left)} Pixel distributions of benchmarks. ObjectNet~\citep{objectnet} contains significantly higer resolution than ImageNet~\citep{imagenet}, which is downscaled. \textbf{(Right)} Accuracy on benchmarks when increasing the pixel count (tokens) used by vision encoder when creating embeddings.}
    \label{fig:resolution}
\end{figure*}

\subsection{Zero-shot Retrieval and Classification}
\label{sec:rep-quality}

We first evaluate the quality of \textbf{ABC} embeddings, as controlling low-quality visual embeddings is useless.
To test embedding quality, we use two zero-shot tasks: retrieval from a pool of candidates and image classification.  
For retrieval, we use MSCOCO~\citep{mscoco} and Flickr30K~\citep{flickr}.
For image classification, we use MMEB~\cite{vlm2vec}, which provides splits of many standard image classification tasks, such as ImageNet-1K~\citep{imagenet}, VOC-2007~\citep{voc-2007}, and ObjectNet~\citep{objectnet}.
We note that the classification labels in MMEB are short, often only one or two words.
This is problematic for image-captioning models that have been trained on full sentences~\citep{coca}.
To alleviate this issue, we use \citet{clip}'s technique of embedding classification labels in a sentence template.
We use ``$\texttt{A photo of a \{label\}.}$'' as our template for all classification evaluations.

In \cref{tab:retrieval_results}, our model demonstrates strong image-to-text retrieval capabilities, achieving competitive performance to models that have been contrastively trained with hundreds of GPUs and massive batch sizes~\citep{eva-clip, open_clip}.
We achieve the best performance in MSCOCO image-to-text retrieval.
Comparatively, our text-to-image retrieval is weaker, as we under-perform both EVA-02-CLIP-E+ and OpenCLIP-G.
%
%
Averaged across all classification tasks (\cref{tab:classification_results}), we achieve $11.2\%$ better accuracy than the next-best model.
On ImageNet-1K~\citep{imagenet}, \textbf{ABC} has $7.7\%$ better zero-shot accuracy than OpenCLIP.
\textbf{ABC}'s performance on established benchmarks such as MSCOCO and ImageNet-1K indicate that the pretrained model's embeddings are useful.
In \cref{sec:control}, we evaluate the fine-tuned adapter's ability to embed images with text instructions into the pretrained model's embedding space.

\subsection{Image Resolution and Embedding Quality}

\begin{table}[!b]
\centering

\resizebox{0.45\columnwidth}{!}{%
\begin{tabular}{lcc}
\toprule
\textbf{\# Negatives} & \textbf{Image $\rightarrow$ Text} & \textbf{Text $\rightarrow$ Image} \\
\midrule
0 & 50.4 & 35.7 \\
3 & 58.4 & 44.0 \\
6 & 59.9 & 44.8 \\
\bottomrule
\end{tabular}%
}
\caption{Top-1 retrieval performance on MSCOCO with varying number of mined hard negatives.}
\label{tab:neg_ablation}
\end{table}

\label{sec:resolution}

Our VLLM backbone, Qwen2-VL-7B~\citep{qwenvl}, supports image inputs with variable resolution.
This allows the user to effectively trade-off image resolution with inference speed by adjusting the number of tokens output by the vision encoder.
We examine how this trade-off influences embedding quality for image classification.
We find that performance on certain tasks, like ObjectNet~\citep{objectnet} and ImageNet-A~\citep{imagenet-a}, strongly correlate with the resolution used in the VLLM vision encoder~(\cref{fig:resolution}).
On average, ObjectNet images have $13$ times more pixels than those from ImageNet-1K.
When the number of tokens produced by the vision encoder is scaled up, we see a large improvement on ObjectNet, with accuracy increasing by 23.4\%.
However, lower resolution benchmarks like ImageNet-1K do not benefit nearly as much from scaling resolution.
Notably, a smartphone camera takes photos with significantly higher resolution than images in ImageNet-1K or ObjectNet, by default~\citep{phonespecs}.
This motivates the need for benchmarks containing larger resolution images.
Existing low-resolution benchmarks may underestimate the capabilities of models that can natively utilize high-resolution images.
Furthermore, we find that correctly classifying the ``natural adversarial examples'' of ImageNet-A~\citep{imagenet-a} is largely a function of resolution.
Our accuracy on ImageNet-A increases from only $12\%$ to $47.5\%$ simply by scaling the resolution used by the vision encoder during evaluation.

\subsection{Mined Negatives and Temperature}
\label{sec:pretrain-loss}

\begin{figure*}[ht]
    \centering
    \begin{minipage}[t]{0.32\textwidth}
        \centering
        \includegraphics[width=\textwidth]{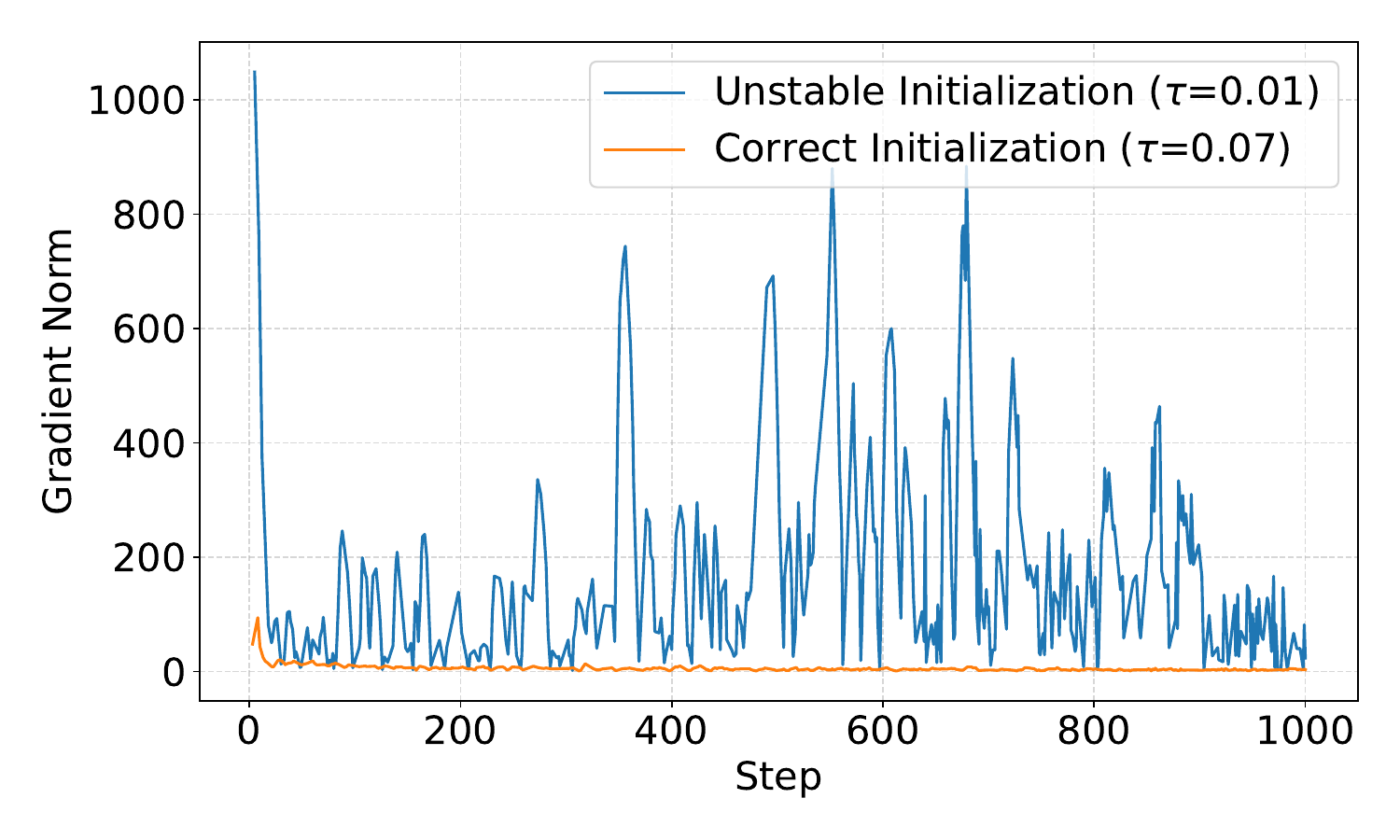}
        \label{fig:temp}
    \end{minipage}
    \hfill
    \begin{minipage}[t]{0.32\textwidth}
        \centering
        \includegraphics[width=\textwidth]{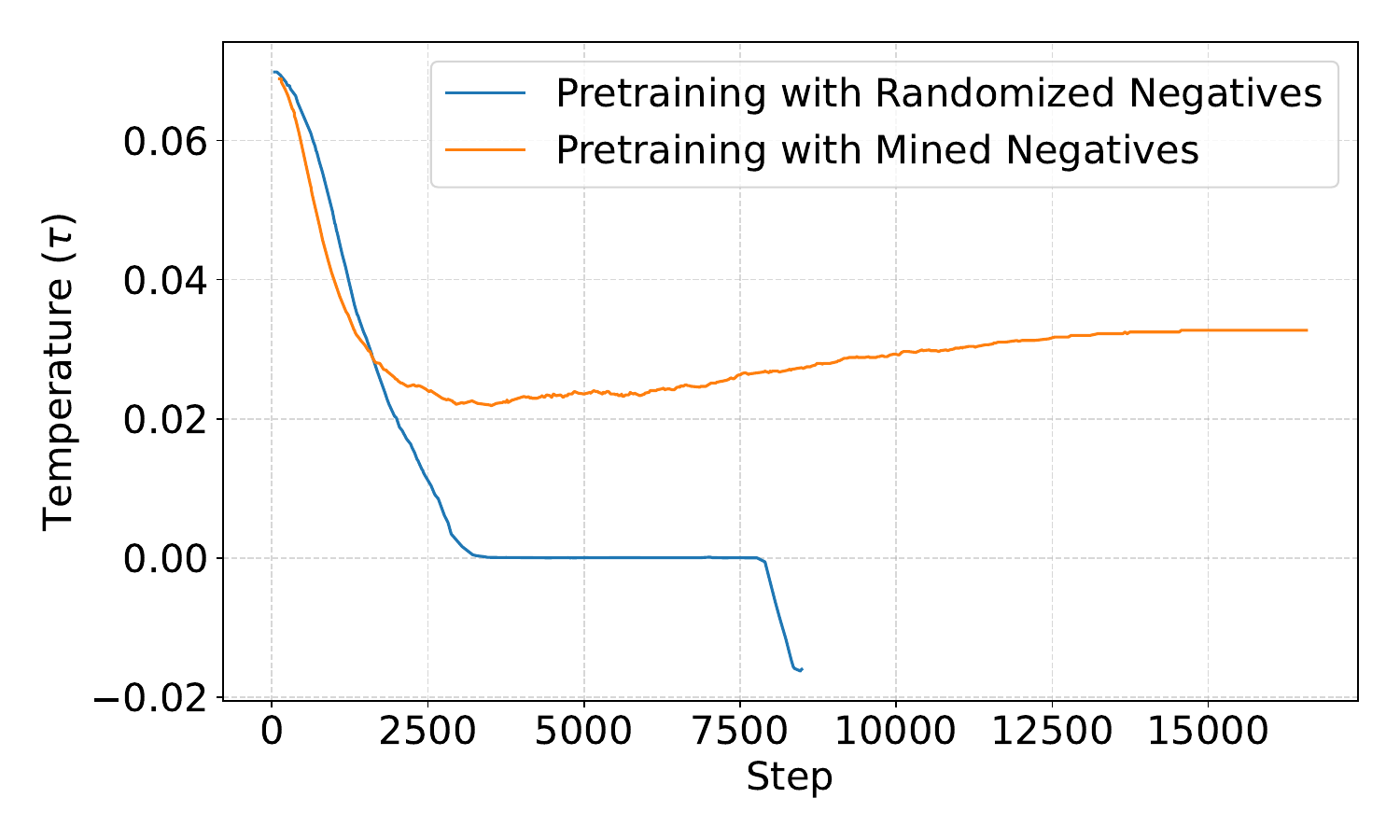}
        \label{fig:temperature_pretraining}
    \end{minipage}
    \hfill
    \begin{minipage}[t]{0.32\textwidth}
        \centering
        \includegraphics[width=\textwidth]{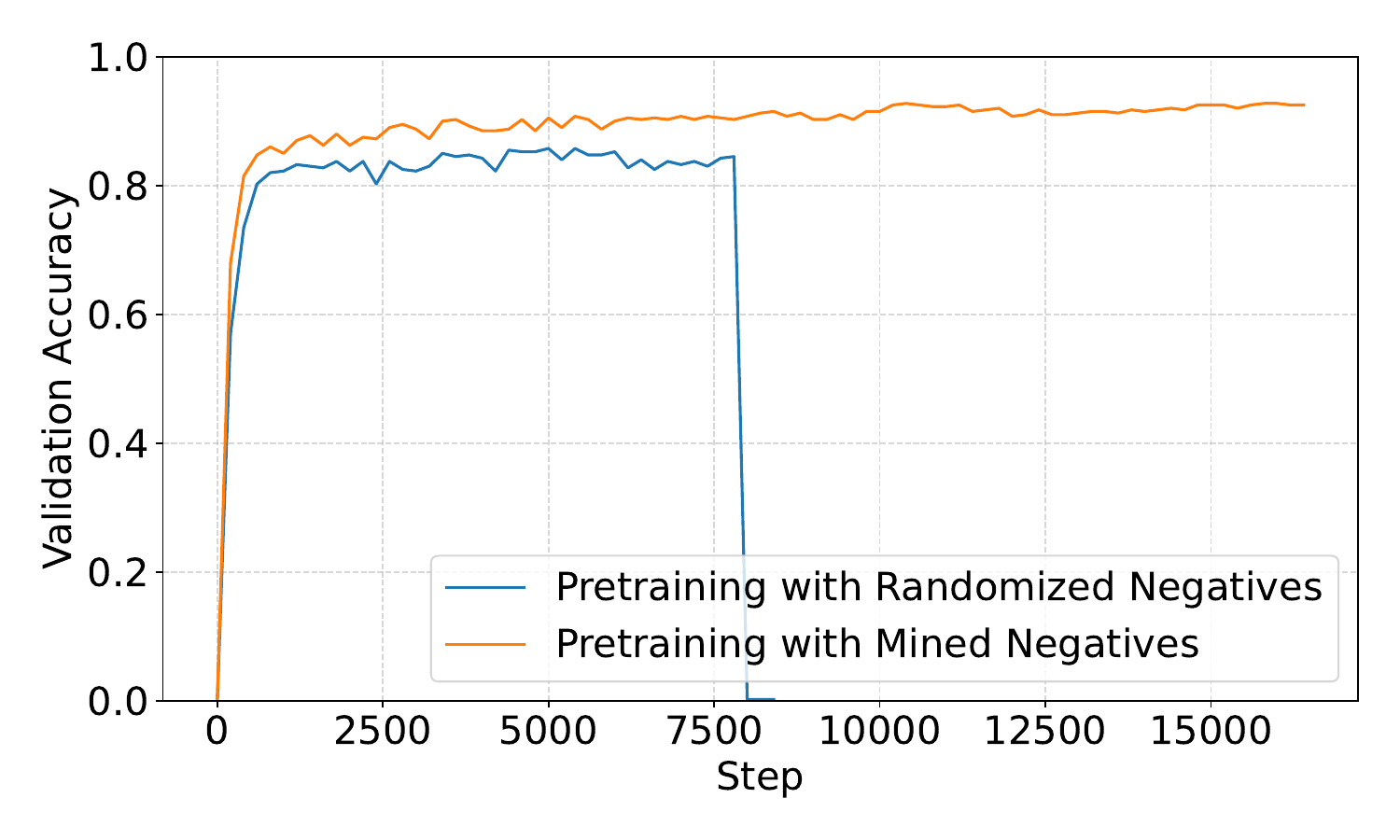}
        \label{fig:acc_pretraining}
    \end{minipage}

    \vspace{-0.5cm}
    \caption{Visualization of temperature ($\tau$) and its effects on training dynamics. \textbf{(Left)} The effect of temperature initializations on gradient norm while training our negative mining model. \textbf{(Middle)} Temperature behavior during pretraining, without and without mined negatives. \textbf{(Right)} Validation accuracy during pretraining, with and without mined negatives.}
    \label{fig:combined_temperature}
\end{figure*}

The setting of the temperature hyperparameter ($\tau$) is known to be crucial for contrastive training~\citep{align}.
However, $\tau$ is often treated as a fixed hyperparameter~\citep{vlm2vec, uniir}.
A large fixed setting of $\tau$ results in a significantly worse model~\citep{wang2021understandingbehaviourcontrastiveloss}, whereas a too small setting can result in training instability (\cref{fig:combined_temperature}).
Therefore, we find that optimizing $\tau$ throughout the pretraining process is crucial.
In shorter runs like training our negative mining model, we find that shows that \emph{both} the initialization and optimization of $\tau$ is crucial.
For our negative mining model, we find $\tau = 0.07$ to be a temperature initialization that is both stable and performant.

\textbf{Mined negatives and temperature.} \revised{In {\cref{tab:neg_ablation}}, we ablate over the number of mined hard negatives used during pretraining. Each pretraining run consists of 500 steps with a batch size of 256.
Introducing 6 mined hard negatives improves top-1 image to text retrieval by $9.5\%$ and top-1 text to image retrieval by $9.1\%$.}
Furthermore, we find that \emph{the mined negatives are essential for training stability}.
\cref{fig:combined_temperature} shows our pretraining run with a batch size of 128 images queries and 1024 text candidates.
\revised{We compare two runs: one using randomized negatives sampled uniformly randomly from our dataset and the other our mined hard negatives.}
We find that if randomized negatives are used instead of our mined negatives, the optimizer tends to push $\tau$ very close to $0$.
This results in numerical instability, loss spikes, and eventually catastrophic failure.
When our mined negatives are used in our pretraining, the temperature stabilizes at a very reasonable value of around $0.03$.

\subsection{VLLM Backbone}
\label{sec:backbone}

We explore how the choice of VLLM backbone has an effect on the quality of our multimodal embedding.
We ablate over $3$ popular choices of VLLM backbone, all at approximately the $8$ billion parameter scale:
{Qwen2-VL-7B}~\citep{qwenvl}, our chosen backbone.
{InternVL-2-8B} from the internVL family of VLMs~\citep{internvl_1}.
Lastly, {LLaVA-NeXT}~\citep{llavanext} with Mistral-7B~\citep{mistral} as its LLM component.
We pretrain each model for $1000$ steps with a query batch size of $128$ and candidate batch size of $1024$.

\begin{table}[H]
\centering
\resizebox{0.5\columnwidth}{!}{%
\begin{tabular}{lcccc}
\toprule
\textbf{Model} & $\text{Accuracy}_{\text{val}}$ & $\text{MMLU}_{\text{val}}$ & $\text{MMBench}_{\text{EN}}$ \\
\midrule
\textbf{LLaVA-NeXT}   & 62.6   & 35.3   & 68.7 \\
\textbf{InternVL-2-8B} & 65.9   & 51.8  & 81.7   \\
\textbf{Qwen2-VL-7B}  & \textbf{70.2}  & \textbf{54.1}   & \textbf{83.0}   \\
\bottomrule
\end{tabular}%
}
\caption{Ablation over VLLM backbone choice.}
\label{tab:backbone}
\end{table}
\vspace{-0.25cm}

\cref{tab:backbone} shows the validation accuracy of \textbf{ABC} with different backbones.
We find that our backbone choice, Qwen2-VL-7B, produces the best results.
We also note each backbone's performance on two standard generative VLLM benchmarks: MMMU~\cite{mmmu} and MMBench~\cite{mmbench}.
We find that performance after contrastive training strongly correlates with the performance of the backbone on generative tasks.
This indicates that training better VLLMs naturally results in better backbones for our model.

\subsection{Controlling Embeddings using Natural Language}
\label{sec:control}

\begin{table*}[!b]
\centering
\resizebox{1\textwidth}{!}{
\begin{tabular}{lcccccccc}
\toprule
& \textbf{CLIP} & \textbf{OpenCLIP} & \textbf{SigLIP} & \textbf{UniIR} & \textbf{MagicLens} & \textbf{BLIP2} & \textbf{MMRet} & \textbf{ABC (ours)} \\
\midrule
\rowcolor{LightGray} \textbf{VQA (10 tasks)} & & & & & & & & \\
OK-VQA~\cite{okvqa}              
& 7.5  & 11.5 & 2.4  & 25.4 & 12.7 & 8.7  & \underline{28.0} & \textbf{48.1} \\
A-OKVQA~\cite{a-okvqa}             
& 3.8  & 3.3  & 1.5  & 8.8  & 2.9  & 3.2  & \underline{11.6} & \textbf{37.3} \\
DocVQA~\cite{docvqa}              
& 4.0  & 5.3  & 4.2  & 6.2  & 3.0  & 2.6  & \underline{12.6} & \textbf{28.5} \\
InfographicsVQA~\cite{infographic}     
& 4.6  & 4.6  & 2.7  & 4.6  & 5.9  & 2.0  & \textbf{10.6} & \underline{7.9} \\
ChartQA~\cite{chartqa}             
& 1.4  & 1.5  & \underline{3.0}  & 1.6  & 0.9  & 0.5  & 2.4  & \textbf{11.7} \\
Visual7W~\cite{visual7w}            
& 4.0  & 2.6  & 1.2  & \underline{14.5} & 2.5  & 1.3  & 9.0  & \textbf{25.6} \\
ScienceQA~\cite{scienceqa}           
& 9.4  & 10.2 & 7.9  & 12.8 & 5.2  & 6.8  & \underline{23.3} & \textbf{26.3} \\
VizWiz~\cite{wizviz}              
& 8.2  & 6.6  & 2.3  & 24.3 & 1.7  & 4.0  & \underline{25.9} & \textbf{29.4} \\
GQA~\cite{gqa}                 
& 41.3 & 52.5 & \underline{57.5} & 48.8 & 43.5 & 9.7  & 41.3 & \textbf{60.1} \\
TextVQA\cite{textvqa}             
& 7.0  & 10.9 & 1.0  & 15.1 & 4.6  & 3.3  & \underline{18.9} & \textbf{35.4} \\
\midrule
\textit{All VQA} 
& 9.1  & 10.9 & 8.4  & 16.2 & 8.3  & 4.2  & \underline{18.4} & \textbf{31.0} \\
\bottomrule
\end{tabular}
}
\caption{Zero-shot VQA results on MMEB~\citep{vlm2vec}. Best is \textbf{bold}, second best is \underline{underlined}.}
\label{tab:vqa_results}
\end{table*}


\Cref{tab:vqa_results} shows \textbf{ABC}'s performance on the VQA split of MMEB.
In just 100 training steps using instructions, our model surpasses MMRet~\citep{megapairs} on VQA, which was trained with millions of instructions.
Interestingly, CLIP derivatives fine-tuned with instructions such as MagicLens~\citep{magiclens} and UniIR~\citep{uniir} do not perform significantly better than their respective baselines, despite MMEB providing instructions for each task.
This evidences that encoder-only embedding models struggle to effectively utilize natural language instructions~\citep{vlm2vec}.

\begin{wrapfigure}{r}{0.35\textwidth}
  \centering
\includegraphics[width=0.35\textwidth]{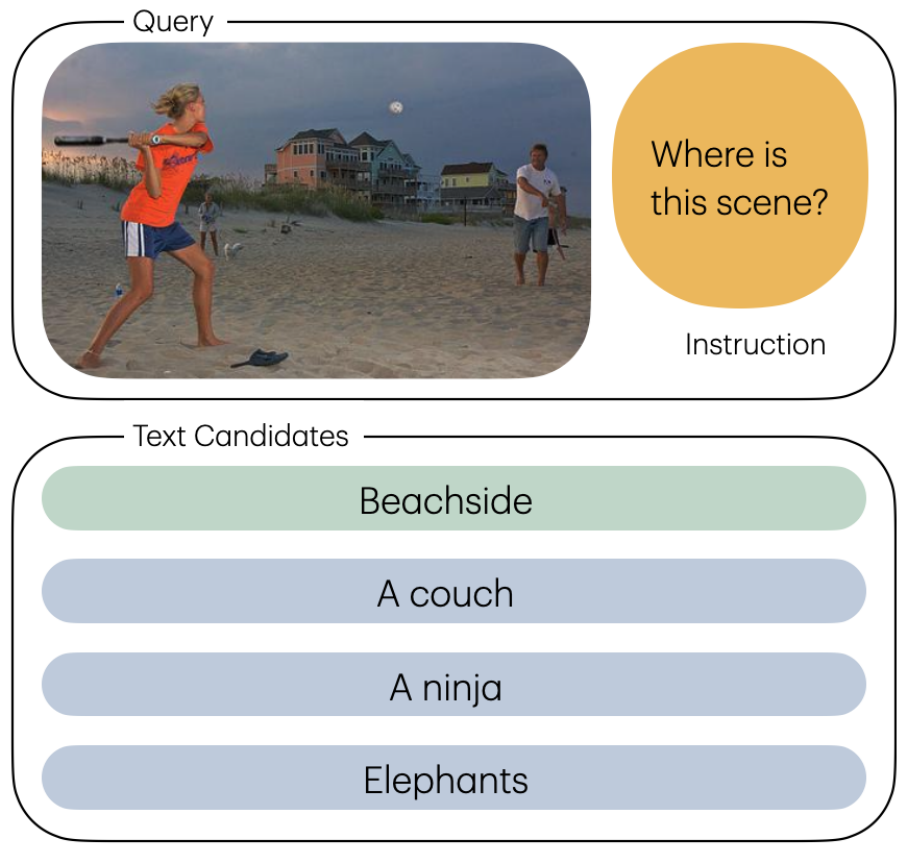}
\caption{An example of a VQA task from the Visual7W~\citep{visual7w} split of MMEB~\citep{vlm2vec}.}
\label{fig:vqa_flaw}
\end{wrapfigure}

\textbf{Problems with VQA.}
A VQA benchmark should require the use of both modalities \emph{together}, and using only the image or the text should be insufficient to accomplish the task.
We note that ensuring this property requires inspecting not only the queries but the candidate pool as well.
Consider \cref{fig:vqa_flaw}, an example from the Visual7W task in MMEB.
When the candidate pool is examined, it is clear that there is only one answer that is plausible, given the image.
Therefore, the instruction is not required to accomplish the VQA task.
This a common pitfall when adapting open-ended generative VQA benchmarks into fixed candidate pool embedding tasks.

\textbf{\texttt{CtrlBench.}}
We construct \texttt{CtrlBench}, an instruction-controlled retrieval benchmark.
\texttt{CtrlBench} has a similar format to the Flicker30K~\citep{flickr} test split.
However, instead of each image having multiple valid captions, we instead provide an instruction from which the model can infer the most relevant text candidate.
Therefore, \texttt{CtrlBench} tests both retrieval and instruction following capabilities.
\revised{To construct \texttt{CtrlBench}, we sample a $1000$ images from ADE20K~\citep{Zhou_2017_CVPR}.
To create $5000$ instructions and text candidate pairs, we generate $5$ instructions for each image, each corresponding to a distinct aspect of the image.
To accomplish this, we use Gemini 2.5 Flash~\citep{geminiteam2025geminifamilyhighlycapable}.
We provide the exact prompt that we use in \Cref{apx:cbenchprompt}.}
As each image has $5$ associated captions in the candidate pool, a model that cannot utilize instructions can (in expectation) achieve at most $20\%$ R@1 on the benchmark.
\revised{Therefore, \texttt{CtrlBench} requires the embedding model to use the provided instructions to disambiguate visual queries.}
We deduplicate captions from the dataset to prevent ambiguity in our text candidates, resulting in \texttt{CtrlBench} being slightly smaller than 1000 images.
%
\begin{table}[H]
\centering
\resizebox{0.5\columnwidth}{!}{%
\begin{tabular}{lccc}
\toprule
\textbf{Model} & R@1 & R@5 & R@10 \\
\midrule
\textbf{UniIR}~\cite{uniir}   & 0.9   & 2.59   & 4.30   \\
\textbf{MagicLens}~\cite{magiclens}  & 17.0   & 37.33   & 48.47   \\
\textbf{VLM2Vec}~\cite{vlm2vec}   & 2.95  & 8.49   & 12.36   \\
\textbf{GME-Qwen2VL-2B}~\cite{gme}   & 20.92  & 46.33   & 57.27   \\
\textbf{GME-Qwen2VL-7B}~\cite{gme}   & 24.75  & 54.48   & 65.27   \\
\rowcolor{LightGray}
\textbf{ABC}~(pretrained)       & {11.71} & {32.48} & {42.69} \\
\rowcolor{LightGray}
\textbf{ABC}~(finetuned)       & \textbf{50.94} & \textbf{78.43} & \textbf{87.47} \\
\bottomrule
\end{tabular}%
}
\caption{Recall@K  on \texttt{CtrlBench}.}
\label{tab:instruction}
\end{table}

\cref{tab:instruction} shows the performance of multimodal embedding models on \texttt{CtrlBench}.
We compare with $2$ instruction fine-tuned CLIP derivatives: UniIR~\cite{uniir} and MagicLens~\cite{magiclens} as well as VLM2Vec~\cite{vlm2vec} and GME~\citep{gme}, two concurrent works that adapt VLLMs into multimodal embedding models.
We find that neither of the CLIP-based architectures have above $20\%$ R@1 on \texttt{CtrlBench}, the performance that indicates that the model is non-trivially utilizing instructions.
We find that VLLM architectures are better at instruction-controlled retrieval.
\textbf{ABC} and VLM2Vec have R@1 above $20\%$, proving that they can non-trivially utilize instructions.

\section{{Discussion and Related Work}}

\subsection{Distribution Overlap in Training and Evaluation}

Throughout our experiments, we found that several training data sources overlap with commonly used benchmarks.
Several benchmark are constructed from the same underlying data source as training data or directly derived from training data~\citep{objectnet,imagenet, flickr, mscoco}.
This makes evaluating ``zero-shot'' capabilities difficult.
To the author's knowledge, our pretraining data~\citep{conceptualcaptions} and instruction-finetuning data~\cite{visualgenome} does not overlap with any benchmarks we evaluate against, and all evaluations (\Cref{tab:retrieval_results,tab:classification_results,tab:vqa_results,tab:instruction}) are out-of-distribution.

\subsection{Decoupling Pretraining and Instruction Fine-tuning}

We find that adapting an instruction fine-tuned VLLM to use instructions is relatively cheap.
Our instruction fine-tuning completes in less than an hour on a couple A100 GPUs.
This allowed us to quickly iterate on our instruction fine-tuning stage, without pretraining from scratch.
Comparatively, training a VLLM to create high-quality embeddings is much more resource intensive. 
Our pretraining run barely fits into the 640 GB of VRAM provided by a single A100 node, and takes several days to complete.
Furthermore, our work indicates that these models could benefit substantially from further scaling~(\cref{apx:scale}).

\subsection{Alternative Approaches}
\revised{MagicLens~\citep{magiclens} and UniIR~\citep{uniir} are multimodal embedding models that fuse CLIP embeddings.
By combining multiple CLIP models, they take image-text pairs and align them with image-text pairs using contrastive loss and instruction finetuning.
Due to late fusion of modalities, these architectures struggle to use instructions to modify representations~\citep{vlm2vec}.
Recently, VLM2Vec~\cite{vlm2vec} and GME~\citep{gme} have also adopted contrastive training on VLLM backbones to produce multimodal embeddings. However, these approaches make extensive use of human-annotated instruction datasets, with hundreds of thousands of labeled examples used during training.
This limits the scalability and adaptability of their method.}

\subsection{Important Factors during Pretraining}
Prior work has largely focused on what architectural adaptations to the make the VLLM to convert it into an embedding model.
These include the attention mask, how the embedding is pooled and adapter architecture~\citep{llm2vec, nvembed, vlm2vec}.
Throughout our experiments, we find that most of these choices are often interchangeable or only produce marginal improvements~(\cref{apx:arch}).
However, we find that many of the crucial factors when training CLIP models are also important when adapting VLMs.
In particular, well-chosen data, batch size and number of samples seen during training are all important factors~(\cref{apx:scale}), just like with CLIP models~\citep{datacomp, open_clip}.

\section{Conclusion} We introduce \textbf{ABC}, a multimodal embedding model that leverages a VLLM backbone to control image representations via natural language instructions.
It achieves strong zero-shot results on a variety of multimodal tasks, spanning retrieval, classification, and VQA.
\textbf{ABC} consists of a multi-stage training process, which isolates the computationally expensive contrastive pretraining from a lightweight
instruction finetuning phase.
We explore what factors are the most crucial when adapting VLMs to output multimodal embeddings.
In particular, we find that vision encoder resolution, contrastive loss temperature and VLLM backbone choice are all important factors.
Lastly, we design \texttt{CtrlBench} to measure our model's ability to use instructions to accomplish subtle natural language-guided retrieval tasks.

\section{Broader Impact Statement} 

\revised{The ability to exhibit greater control over visual representations, especially those used information retrieval systems, has potential for both positive and negative impacts.
On the positive side, it allows users greater control over visual information retrieval.
This capability is particularly useful for visually-impaired users that can struggle to directly use and interpret images.
However, introducing natural language control over visual representations has potentially negative impacts.
We highlight two potential harms: {(1)} Accidental introduction of bias due to natural language conditioning.
Large language models exhibit similar biases to humans~\citep{guo2024biaslargelanguagemodels}.
LLM representations used in information retrieval may adopt these biases.
Furthermore, allowing users natural language control over visual representations may result in users accidentally introducing their own biases into visual retrieval.
{(2)} Malicious control over information retrieval. Introducing natural language control over visual representations creates an entry-point to filter or modify what information is retrieved by downstream users. This is alarming in the internet age, where control over information access is a powerful tool for authoritarian control.}

\bibliography{main}

\begin{thebibliography}{69}
\providecommand{\natexlab}[1]{#1}
\providecommand{\url}[1]{\texttt{#1}}
\expandafter\ifx\csname urlstyle\endcsname\relax
  \providecommand{\doi}[1]{doi: #1}\else
  \providecommand{\doi}{doi: \begingroup \urlstyle{rm}\Url}\fi

\bibitem[Aliannejadi et~al.(2019)Aliannejadi, Zamani, Crestani, and Croft]{Aliannejadi_2019}
Mohammad Aliannejadi, Hamed Zamani, Fabio Crestani, and W.~Bruce Croft.
\newblock Asking clarifying questions in open-domain information-seeking conversations.
\newblock In \emph{Proceedings of the 42nd International ACM SIGIR Conference on Research and Development in Information Retrieval}, SIGIR ’19, pp.\  475–484. ACM, July 2019.
\newblock \doi{10.1145/3331184.3331265}.
\newblock URL \url{http://dx.doi.org/10.1145/3331184.3331265}.

\bibitem[Amiraz et~al.(2025)Amiraz, Cuconasu, Filice, and Karnin]{distract}
Chen Amiraz, Florin Cuconasu, Simone Filice, and Zohar Karnin.
\newblock The distracting effect: Understanding irrelevant passages in rag, 2025.
\newblock URL \url{https://arxiv.org/abs/2505.06914}.

\bibitem[{Apple Inc.}(2024)]{phonespecs}
{Apple Inc.}
\newblock iphone 16 -- tech specs.
\newblock \url{https://support.apple.com/en-ca/121029}, 2024.
\newblock Accessed: 2025-01-22.

\bibitem[Barbu et~al.(2019)Barbu, Mayo, Alverio, Luo, Wang, Gutfreund, Tenenbaum, and Katz]{objectnet}
Andrei Barbu, David Mayo, Julian Alverio, William Luo, Christopher Wang, Dan Gutfreund, Josh Tenenbaum, and Boris Katz.
\newblock Objectnet: A large-scale bias-controlled dataset for pushing the limits of object recognition models.
\newblock In H.~Wallach, H.~Larochelle, A.~Beygelzimer, F.~d\textquotesingle Alch\'{e}-Buc, E.~Fox, and R.~Garnett (eds.), \emph{Advances in Neural Information Processing Systems}, volume~32. Curran Associates, Inc., 2019.
\newblock URL \url{https://proceedings.neurips.cc/paper_files/paper/2019/file/97af07a14cacba681feacf3012730892-Paper.pdf}.

\bibitem[BehnamGhader et~al.(2024)BehnamGhader, Adlakha, Mosbach, Bahdanau, Chapados, and Reddy]{llm2vec}
Parishad BehnamGhader, Vaibhav Adlakha, Marius Mosbach, Dzmitry Bahdanau, Nicolas Chapados, and Siva Reddy.
\newblock Llm2vec: Large language models are secretly powerful text encoders, 2024.
\newblock URL \url{https://arxiv.org/abs/2404.05961}.

\bibitem[Changpinyo et~al.(2021)Changpinyo, Sharma, Ding, and Soricut]{c12m}
Soravit Changpinyo, Piyush Sharma, Nan Ding, and Radu Soricut.
\newblock Conceptual 12m: Pushing web-scale image-text pre-training to recognize long-tail visual concepts.
\newblock \emph{CoRR}, abs/2102.08981, 2021.
\newblock URL \url{https://arxiv.org/abs/2102.08981}.

\bibitem[Chen et~al.(2024)Chen, Wu, Wang, Su, Chen, Xing, Zhong, Zhang, Zhu, Lu, Li, Luo, Lu, Qiao, and Dai]{internvl_1}
Zhe Chen, Jiannan Wu, Wenhai Wang, Weijie Su, Guo Chen, Sen Xing, Muyan Zhong, Qinglong Zhang, Xizhou Zhu, Lewei Lu, Bin Li, Ping Luo, Tong Lu, Yu~Qiao, and Jifeng Dai.
\newblock Internvl: Scaling up vision foundation models and aligning for generic visual-linguistic tasks, 2024.
\newblock URL \url{https://arxiv.org/abs/2312.14238}.

\bibitem[Cherti et~al.(2023{\natexlab{a}})Cherti, Beaumont, Wightman, Wortsman, Ilharco, Gordon, Schuhmann, Schmidt, and Jitsev]{open_clip}
Mehdi Cherti, Romain Beaumont, Ross Wightman, Mitchell Wortsman, Gabriel Ilharco, Cade Gordon, Christoph Schuhmann, Ludwig Schmidt, and Jenia Jitsev.
\newblock Reproducible scaling laws for contrastive language-image learning.
\newblock In \emph{2023 IEEE/CVF Conference on Computer Vision and Pattern Recognition (CVPR)}. IEEE, June 2023{\natexlab{a}}.
\newblock \doi{10.1109/cvpr52729.2023.00276}.
\newblock URL \url{http://dx.doi.org/10.1109/CVPR52729.2023.00276}.

\bibitem[Cherti et~al.(2023{\natexlab{b}})Cherti, Beaumont, Wightman, Wortsman, Ilharco, Gordon, Schuhmann, Schmidt, and Jitsev]{scaling_laws}
Mehdi Cherti, Romain Beaumont, Ross Wightman, Mitchell Wortsman, Gabriel Ilharco, Cade Gordon, Christoph Schuhmann, Ludwig Schmidt, and Jenia Jitsev.
\newblock Reproducible scaling laws for contrastive language-image learning.
\newblock In \emph{2023 IEEE/CVF Conference on Computer Vision and Pattern Recognition (CVPR)}, pp.\  2818–2829. IEEE, June 2023{\natexlab{b}}.
\newblock \doi{10.1109/cvpr52729.2023.00276}.
\newblock URL \url{http://dx.doi.org/10.1109/CVPR52729.2023.00276}.

\bibitem[Djolonga et~al.(2020)Djolonga, Yung, Tschannen, Romijnders, Beyer, Kolesnikov, Puigcerver, Minderer, D'Amour, Moldovan, Gelly, Houlsby, Zhai, and Lucic]{imagenet-a}
Josip Djolonga, Jessica Yung, Michael Tschannen, Rob Romijnders, Lucas Beyer, Alexander Kolesnikov, Joan Puigcerver, Matthias Minderer, Alexander D'Amour, Dan Moldovan, Sylvain Gelly, Neil Houlsby, Xiaohua Zhai, and Mario Lucic.
\newblock On robustness and transferability of convolutional neural networks.
\newblock \emph{CoRR}, abs/2007.08558, 2020.
\newblock URL \url{https://arxiv.org/abs/2007.08558}.

\bibitem[Everingham et~al.()Everingham, Van~Gool, Williams, Winn, and Zisserman]{voc-2007}
M.~Everingham, L.~Van~Gool, C.~K.~I. Williams, J.~Winn, and A.~Zisserman.
\newblock The {PASCAL} {V}isual {O}bject {C}lasses {C}hallenge 2007 {(VOC2007)} {R}esults.
\newblock http://www.pascal-network.org/challenges/VOC/voc2007/workshop/index.html.
\newblock URL \url{"http://host.robots.ox.ac.uk/pascal/VOC/voc2007"}.

\bibitem[Gadre et~al.(2023)Gadre, Ilharco, Fang, Hayase, Smyrnis, Nguyen, Marten, Wortsman, Ghosh, Zhang, Orgad, Entezari, Daras, Pratt, Ramanujan, Bitton, Marathe, Mussmann, Vencu, Cherti, Krishna, Koh, Saukh, Ratner, Song, Hajishirzi, Farhadi, Beaumont, Oh, Dimakis, Jitsev, Carmon, Shankar, and Schmidt]{datacomp}
Samir~Yitzhak Gadre, Gabriel Ilharco, Alex Fang, Jonathan Hayase, Georgios Smyrnis, Thao Nguyen, Ryan Marten, Mitchell Wortsman, Dhruba Ghosh, Jieyu Zhang, Eyal Orgad, Rahim Entezari, Giannis Daras, Sarah Pratt, Vivek Ramanujan, Yonatan Bitton, Kalyani Marathe, Stephen Mussmann, Richard Vencu, Mehdi Cherti, Ranjay Krishna, Pang~Wei Koh, Olga Saukh, Alexander Ratner, Shuran Song, Hannaneh Hajishirzi, Ali Farhadi, Romain Beaumont, Sewoong Oh, Alex Dimakis, Jenia Jitsev, Yair Carmon, Vaishaal Shankar, and Ludwig Schmidt.
\newblock Datacomp: In search of the next generation of multimodal datasets, 2023.
\newblock URL \url{https://arxiv.org/abs/2304.14108}.

\bibitem[Gao et~al.(2021)Gao, Zhang, Han, and Callan]{gradcache}
Luyu Gao, Yunyi Zhang, Jiawei Han, and Jamie Callan.
\newblock Scaling deep contrastive learning batch size under memory limited setup, 2021.
\newblock URL \url{https://arxiv.org/abs/2101.06983}.

\bibitem[Gudibande et~al.(2023)Gudibande, Wallace, Snell, Geng, Liu, Abbeel, Levine, and Song]{ins_overfit1}
Arnav Gudibande, Eric Wallace, Charlie Snell, Xinyang Geng, Hao Liu, Pieter Abbeel, Sergey Levine, and Dawn Song.
\newblock The false promise of imitating proprietary llms, 2023.
\newblock URL \url{https://arxiv.org/abs/2305.15717}.

\bibitem[Guo et~al.(2024)Guo, Guo, Su, Yang, Zhu, Li, Qiu, and Liu]{guo2024biaslargelanguagemodels}
Yufei Guo, Muzhe Guo, Juntao Su, Zhou Yang, Mengqiu Zhu, Hongfei Li, Mengyang Qiu, and Shuo~Shuo Liu.
\newblock Bias in large language models: Origin, evaluation, and mitigation, 2024.
\newblock URL \url{https://arxiv.org/abs/2411.10915}.

\bibitem[Gurari et~al.(2018)Gurari, Li, Stangl, Guo, Lin, Grauman, Luo, and Bigham]{wizviz}
Danna Gurari, Qing Li, Abigale~J. Stangl, Anhong Guo, Chi Lin, Kristen Grauman, Jiebo Luo, and Jeffrey~P. Bigham.
\newblock Vizwiz grand challenge: Answering visual questions from blind people, 2018.
\newblock URL \url{https://arxiv.org/abs/1802.08218}.

\bibitem[Hadsell et~al.(2006)Hadsell, Chopra, and LeCun]{contrastive_loss}
R.~Hadsell, S.~Chopra, and Y.~LeCun.
\newblock Dimensionality reduction by learning an invariant mapping.
\newblock In \emph{2006 IEEE Computer Society Conference on Computer Vision and Pattern Recognition (CVPR'06)}, volume~2, pp.\  1735--1742, 2006.
\newblock \doi{10.1109/CVPR.2006.100}.

\bibitem[Hendrycks et~al.(2021)Hendrycks, Basart, Mu, Kadavath, Wang, Dorundo, Desai, Zhu, Parajuli, Guo, Song, Steinhardt, and Gilmer]{imagenet-r}
Dan Hendrycks, Steven Basart, Norman Mu, Saurav Kadavath, Frank Wang, Evan Dorundo, Rahul Desai, Tyler Zhu, Samyak Parajuli, Mike Guo, Dawn Song, Jacob Steinhardt, and Justin Gilmer.
\newblock The many faces of robustness: A critical analysis of out-of-distribution generalization, 2021.
\newblock URL \url{https://arxiv.org/abs/2006.16241}.

\bibitem[Hu et~al.(2021)Hu, Shen, Wallis, Allen-Zhu, Li, Wang, Wang, and Chen]{lora}
Edward~J. Hu, Yelong Shen, Phillip Wallis, Zeyuan Allen-Zhu, Yuanzhi Li, Shean Wang, Lu~Wang, and Weizhu Chen.
\newblock Lora: Low-rank adaptation of large language models, 2021.
\newblock URL \url{https://arxiv.org/abs/2106.09685}.

\bibitem[Hudson \& Manning(2019)Hudson and Manning]{gqa}
Drew~A. Hudson and Christopher~D. Manning.
\newblock Gqa: A new dataset for real-world visual reasoning and compositional question answering, 2019.
\newblock URL \url{https://arxiv.org/abs/1902.09506}.

\bibitem[Jia et~al.(2021)Jia, Yang, Xia, Chen, Parekh, Pham, Le, Sung, Li, and Duerig]{align}
Chao Jia, Yinfei Yang, Ye~Xia, Yi-Ting Chen, Zarana Parekh, Hieu Pham, Quoc Le, Yun-Hsuan Sung, Zhen Li, and Tom Duerig.
\newblock Scaling up visual and vision-language representation learning with noisy text supervision.
\newblock In Marina Meila and Tong Zhang (eds.), \emph{Proceedings of the 38th International Conference on Machine Learning}, volume 139 of \emph{Proceedings of Machine Learning Research}, pp.\  4904--4916. PMLR, 18--24 Jul 2021.
\newblock URL \url{https://proceedings.mlr.press/v139/jia21b.html}.

\bibitem[Jiang et~al.(2023)Jiang, Sablayrolles, Mensch, Bamford, Chaplot, de~las Casas, Bressand, Lengyel, Lample, Saulnier, Lavaud, Lachaux, Stock, Scao, Lavril, Wang, Lacroix, and Sayed]{mistral}
Albert~Q. Jiang, Alexandre Sablayrolles, Arthur Mensch, Chris Bamford, Devendra~Singh Chaplot, Diego de~las Casas, Florian Bressand, Gianna Lengyel, Guillaume Lample, Lucile Saulnier, Lélio~Renard Lavaud, Marie-Anne Lachaux, Pierre Stock, Teven~Le Scao, Thibaut Lavril, Thomas Wang, Timothée Lacroix, and William~El Sayed.
\newblock Mistral 7b, 2023.
\newblock URL \url{https://arxiv.org/abs/2310.06825}.

\bibitem[Jiang et~al.(2024)Jiang, Meng, Yang, Yavuz, Zhou, and Chen]{vlm2vec}
Ziyan Jiang, Rui Meng, Xinyi Yang, Semih Yavuz, Yingbo Zhou, and Wenhu Chen.
\newblock Vlm2vec: Training vision-language models for massive multimodal embedding tasks, 2024.
\newblock URL \url{https://arxiv.org/abs/2410.05160}.

\bibitem[Keyvan \& Huang(2022)Keyvan and Huang]{10.1145/3534965}
Kimiya Keyvan and Jimmy~Xiangji Huang.
\newblock How to approach ambiguous queries in conversational search: A survey of techniques, approaches, tools, and challenges.
\newblock \emph{ACM Comput. Surv.}, 55\penalty0 (6), December 2022.
\newblock ISSN 0360-0300.
\newblock \doi{10.1145/3534965}.
\newblock URL \url{https://doi.org/10.1145/3534965}.

\bibitem[Kiela et~al.(2021)Kiela, Firooz, Mohan, Goswami, Singh, Fitzpatrick, Bull, Lipstein, Nelli, Zhu, Muennighoff, Velioglu, Rose, Lippe, Holla, Chandra, Rajamanickam, Antoniou, Shutova, Yannakoudakis, Sandulescu, Ozertem, Pantel, Specia, and Parikh]{hateful-memes}
Douwe Kiela, Hamed Firooz, Aravind Mohan, Vedanuj Goswami, Amanpreet Singh, Casey~A. Fitzpatrick, Peter Bull, Greg Lipstein, Tony Nelli, Ron Zhu, Niklas Muennighoff, Riza Velioglu, Jewgeni Rose, Phillip Lippe, Nithin Holla, Shantanu Chandra, Santhosh Rajamanickam, Georgios Antoniou, Ekaterina Shutova, Helen Yannakoudakis, Vlad Sandulescu, Umut Ozertem, Patrick Pantel, Lucia Specia, and Devi Parikh.
\newblock The hateful memes challenge: Competition report.
\newblock In Hugo~Jair Escalante and Katja Hofmann (eds.), \emph{Proceedings of the NeurIPS 2020 Competition and Demonstration Track}, volume 133 of \emph{Proceedings of Machine Learning Research}, pp.\  344--360. PMLR, 06--12 Dec 2021.
\newblock URL \url{https://proceedings.mlr.press/v133/kiela21a.html}.

\bibitem[Kim et~al.(2023)Kim, Kim, Jeon, Park, and Kang]{kim2023treeclarificationsansweringambiguous}
Gangwoo Kim, Sungdong Kim, Byeongguk Jeon, Joonsuk Park, and Jaewoo Kang.
\newblock Tree of clarifications: Answering ambiguous questions with retrieval-augmented large language models, 2023.
\newblock URL \url{https://arxiv.org/abs/2310.14696}.

\bibitem[Klambauer et~al.(2017)Klambauer, Unterthiner, Mayr, and Hochreiter]{selu}
G\"{u}nter Klambauer, Thomas Unterthiner, Andreas Mayr, and Sepp Hochreiter.
\newblock Self-normalizing neural networks.
\newblock In I.~Guyon, U.~Von Luxburg, S.~Bengio, H.~Wallach, R.~Fergus, S.~Vishwanathan, and R.~Garnett (eds.), \emph{Advances in Neural Information Processing Systems}, volume~30. Curran Associates, Inc., 2017.
\newblock URL \url{https://proceedings.neurips.cc/paper_files/paper/2017/file/5d44ee6f2c3f71b73125876103c8f6c4-Paper.pdf}.

\bibitem[Krishna et~al.(2016)Krishna, Zhu, Groth, Johnson, Hata, Kravitz, Chen, Kalantidis, Li, Shamma, Bernstein, and Li]{visualgenome}
Ranjay Krishna, Yuke Zhu, Oliver Groth, Justin Johnson, Kenji Hata, Joshua Kravitz, Stephanie Chen, Yannis Kalantidis, Li-Jia Li, David~A. Shamma, Michael~S. Bernstein, and Fei-Fei Li.
\newblock Visual genome: Connecting language and vision using crowdsourced dense image annotations, 2016.
\newblock URL \url{https://arxiv.org/abs/1602.07332}.

\bibitem[Lee et~al.(2024)Lee, Roy, Xu, Raiman, Shoeybi, Catanzaro, and Ping]{nvembed}
Chankyu Lee, Rajarshi Roy, Mengyao Xu, Jonathan Raiman, Mohammad Shoeybi, Bryan Catanzaro, and Wei Ping.
\newblock Nv-embed: Improved techniques for training llms as generalist embedding models, 2024.
\newblock URL \url{https://arxiv.org/abs/2405.17428}.

\bibitem[Li et~al.(2023)Li, Li, Savarese, and Hoi]{blip2}
Junnan Li, Dongxu Li, Silvio Savarese, and Steven Hoi.
\newblock Blip-2: Bootstrapping language-image pre-training with frozen image encoders and large language models, 2023.
\newblock URL \url{https://arxiv.org/abs/2301.12597}.

\bibitem[Lin et~al.(2015)Lin, Maire, Belongie, Bourdev, Girshick, Hays, Perona, Ramanan, Zitnick, and Dollár]{mscoco}
Tsung-Yi Lin, Michael Maire, Serge Belongie, Lubomir Bourdev, Ross Girshick, James Hays, Pietro Perona, Deva Ramanan, C.~Lawrence Zitnick, and Piotr Dollár.
\newblock Microsoft coco: Common objects in context, 2015.
\newblock URL \url{https://arxiv.org/abs/1405.0312}.

\bibitem[Liu et~al.(2024{\natexlab{a}})Liu, Li, Li, and Lee]{llavanext}
Haotian Liu, Chunyuan Li, Yuheng Li, and Yong~Jae Lee.
\newblock Improved baselines with visual instruction tuning, 2024{\natexlab{a}}.
\newblock URL \url{https://arxiv.org/abs/2310.03744}.

\bibitem[Liu et~al.(2024{\natexlab{b}})Liu, Duan, Zhang, Li, Zhang, Zhao, Yuan, Wang, He, Liu, Chen, and Lin]{mmbench}
Yuan Liu, Haodong Duan, Yuanhan Zhang, Bo~Li, Songyang Zhang, Wangbo Zhao, Yike Yuan, Jiaqi Wang, Conghui He, Ziwei Liu, Kai Chen, and Dahua Lin.
\newblock Mmbench: Is your multi-modal model an all-around player?, 2024{\natexlab{b}}.
\newblock URL \url{https://arxiv.org/abs/2307.06281}.

\bibitem[Loshchilov \& Hutter(2019)Loshchilov and Hutter]{adamw}
Ilya Loshchilov and Frank Hutter.
\newblock Decoupled weight decay regularization.
\newblock In \emph{International Conference on Learning Representations}, 2019.
\newblock URL \url{https://openreview.net/forum?id=Bkg6RiCqY7}.

\bibitem[Lu et~al.(2022)Lu, Mishra, Xia, Qiu, Chang, Zhu, Tafjord, Clark, and Kalyan]{scienceqa}
Pan Lu, Swaroop Mishra, Tony Xia, Liang Qiu, Kai-Wei Chang, Song-Chun Zhu, Oyvind Tafjord, Peter Clark, and Ashwin Kalyan.
\newblock Learn to explain: Multimodal reasoning via thought chains for science question answering.
\newblock In \emph{The 36th Conference on Neural Information Processing Systems (NeurIPS)}, 2022.

\bibitem[López-Cifuentes et~al.(2020)López-Cifuentes, Escudero-Viñolo, Bescós, and Álvaro García-Martín]{place365}
Alejandro López-Cifuentes, Marcos Escudero-Viñolo, Jesús Bescós, and Álvaro García-Martín.
\newblock Semantic-aware scene recognition.
\newblock \emph{Pattern Recognition}, 102:\penalty0 107256, 2020.
\newblock ISSN 0031-3203.
\newblock \doi{https://doi.org/10.1016/j.patcog.2020.107256}.
\newblock URL \url{https://www.sciencedirect.com/science/article/pii/S0031320320300613}.

\bibitem[Marino et~al.(2019)Marino, Rastegari, Farhadi, and Mottaghi]{okvqa}
Kenneth Marino, Mohammad Rastegari, Ali Farhadi, and Roozbeh Mottaghi.
\newblock Ok-vqa: A visual question answering benchmark requiring external knowledge, 2019.
\newblock URL \url{https://arxiv.org/abs/1906.00067}.

\bibitem[Masry et~al.(2022)Masry, Long, Tan, Joty, and Hoque]{chartqa}
Ahmed Masry, Do~Xuan Long, Jia~Qing Tan, Shafiq Joty, and Enamul Hoque.
\newblock Chartqa: A benchmark for question answering about charts with visual and logical reasoning, 2022.
\newblock URL \url{https://arxiv.org/abs/2203.10244}.

\bibitem[Mathew et~al.(2021{\natexlab{a}})Mathew, Bagal, Tito, Karatzas, Valveny, and Jawahar]{infographic}
Minesh Mathew, Viraj Bagal, Rubèn~Pérez Tito, Dimosthenis Karatzas, Ernest Valveny, and C.~V Jawahar.
\newblock Infographicvqa, 2021{\natexlab{a}}.
\newblock URL \url{https://arxiv.org/abs/2104.12756}.

\bibitem[Mathew et~al.(2021{\natexlab{b}})Mathew, Karatzas, and Jawahar]{docvqa}
Minesh Mathew, Dimosthenis Karatzas, and C.~V. Jawahar.
\newblock Docvqa: A dataset for vqa on document images, 2021{\natexlab{b}}.
\newblock URL \url{https://arxiv.org/abs/2007.00398}.

\bibitem[Min et~al.(2020)Min, Michael, Hajishirzi, and Zettlemoyer]{min2020ambigqaansweringambiguousopendomain}
Sewon Min, Julian Michael, Hannaneh Hajishirzi, and Luke Zettlemoyer.
\newblock Ambigqa: Answering ambiguous open-domain questions, 2020.
\newblock URL \url{https://arxiv.org/abs/2004.10645}.

\bibitem[Muennighoff et~al.(2023)Muennighoff, Tazi, Magne, and Reimers]{mteb}
Niklas Muennighoff, Nouamane Tazi, Loïc Magne, and Nils Reimers.
\newblock Mteb: Massive text embedding benchmark, 2023.
\newblock URL \url{https://arxiv.org/abs/2210.07316}.

\bibitem[OpenAI(2024)]{gpt4o}
OpenAI.
\newblock Gpt-4o system card, 2024.
\newblock URL \url{https://arxiv.org/abs/2410.21276}.

\bibitem[Plummer et~al.(2016)Plummer, Wang, Cervantes, Caicedo, Hockenmaier, and Lazebnik]{flickr}
Bryan~A. Plummer, Liwei Wang, Chris~M. Cervantes, Juan~C. Caicedo, Julia Hockenmaier, and Svetlana Lazebnik.
\newblock Flickr30k entities: Collecting region-to-phrase correspondences for richer image-to-sentence models, 2016.
\newblock URL \url{https://arxiv.org/abs/1505.04870}.

\bibitem[Qu et~al.(2021)Qu, Ding, Liu, Liu, Ren, Zhao, Dong, Wu, and Wang]{batchgather}
Yingqi Qu, Yuchen Ding, Jing Liu, Kai Liu, Ruiyang Ren, Wayne~Xin Zhao, Daxiang Dong, Hua Wu, and Haifeng Wang.
\newblock {R}ocket{QA}: An optimized training approach to dense passage retrieval for open-domain question answering.
\newblock In Kristina Toutanova, Anna Rumshisky, Luke Zettlemoyer, Dilek Hakkani-Tur, Iz~Beltagy, Steven Bethard, Ryan Cotterell, Tanmoy Chakraborty, and Yichao Zhou (eds.), \emph{Proceedings of the 2021 Conference of the North American Chapter of the Association for Computational Linguistics: Human Language Technologies}, pp.\  5835--5847, Online, June 2021. Association for Computational Linguistics.
\newblock \doi{10.18653/v1/2021.naacl-main.466}.
\newblock URL \url{https://aclanthology.org/2021.naacl-main.466/}.

\bibitem[Radford et~al.(2021)Radford, Kim, Hallacy, Ramesh, Goh, Agarwal, Sastry, Askell, Mishkin, Clark, Krueger, and Sutskever]{clip}
Alec Radford, Jong~Wook Kim, Chris Hallacy, Aditya Ramesh, Gabriel Goh, Sandhini Agarwal, Girish Sastry, Amanda Askell, Pamela Mishkin, Jack Clark, Gretchen Krueger, and Ilya Sutskever.
\newblock Learning transferable visual models from natural language supervision, 2021.
\newblock URL \url{https://arxiv.org/abs/2103.00020}.

\bibitem[Russakovsky et~al.(2015)Russakovsky, Deng, Su, Krause, Satheesh, Ma, Huang, Karpathy, Khosla, Bernstein, Berg, and Fei-Fei]{imagenet}
Olga Russakovsky, Jia Deng, Hao Su, Jonathan Krause, Sanjeev Satheesh, Sean Ma, Zhiheng Huang, Andrej Karpathy, Aditya Khosla, Michael Bernstein, Alexander~C. Berg, and Li~Fei-Fei.
\newblock Imagenet large scale visual recognition challenge, 2015.
\newblock URL \url{https://arxiv.org/abs/1409.0575}.

\bibitem[Schuhmann et~al.(2022)Schuhmann, Beaumont, Vencu, Gordon, Wightman, Cherti, Coombes, Katta, Mullis, Wortsman, Schramowski, Kundurthy, Crowson, Schmidt, Kaczmarczyk, and Jitsev]{laion}
Christoph Schuhmann, Romain Beaumont, Richard Vencu, Cade Gordon, Ross Wightman, Mehdi Cherti, Theo Coombes, Aarush Katta, Clayton Mullis, Mitchell Wortsman, Patrick Schramowski, Srivatsa Kundurthy, Katherine Crowson, Ludwig Schmidt, Robert Kaczmarczyk, and Jenia Jitsev.
\newblock Laion-5b: An open large-scale dataset for training next generation image-text models, 2022.
\newblock URL \url{https://arxiv.org/abs/2210.08402}.

\bibitem[Schwenk et~al.(2022)Schwenk, Khandelwal, Clark, Marino, and Mottaghi]{a-okvqa}
Dustin Schwenk, Apoorv Khandelwal, Christopher Clark, Kenneth Marino, and Roozbeh Mottaghi.
\newblock A-okvqa: A benchmark for visual question answering using world knowledge, 2022.
\newblock URL \url{https://arxiv.org/abs/2206.01718}.

\bibitem[Sharma et~al.(2018)Sharma, Ding, Goodman, and Soricut]{conceptualcaptions}
Piyush Sharma, Nan Ding, Sebastian Goodman, and Radu Soricut.
\newblock Conceptual captions: A cleaned, hypernymed, image alt-text dataset for automatic image captioning.
\newblock In \emph{Annual Meeting of the Association for Computational Linguistics}, 2018.
\newblock URL \url{https://api.semanticscholar.org/CorpusID:51876975}.

\bibitem[Singh et~al.(2019)Singh, Natarajan, Shah, Jiang, Chen, Batra, Parikh, and Rohrbach]{textvqa}
Amanpreet Singh, Vivek Natarajan, Meet Shah, Yu~Jiang, Xinlei Chen, Dhruv Batra, Devi Parikh, and Marcus Rohrbach.
\newblock Towards vqa models that can read, 2019.
\newblock URL \url{https://arxiv.org/abs/1904.08920}.

\bibitem[Singh et~al.(2022)Singh, Hu, Goswami, Couairon, Galuba, Rohrbach, and Kiela]{flava}
Amanpreet Singh, Ronghang Hu, Vedanuj Goswami, Guillaume Couairon, Wojciech Galuba, Marcus Rohrbach, and Douwe Kiela.
\newblock Flava: A foundational language and vision alignment model, 2022.
\newblock URL \url{https://arxiv.org/abs/2112.04482}.

\bibitem[Stelmakh et~al.(2022)Stelmakh, Luan, Dhingra, and Chang]{stelmakh-etal-2022-asqa}
Ivan Stelmakh, Yi~Luan, Bhuwan Dhingra, and Ming-Wei Chang.
\newblock {ASQA}: Factoid questions meet long-form answers.
\newblock In Yoav Goldberg, Zornitsa Kozareva, and Yue Zhang (eds.), \emph{Proceedings of the 2022 Conference on Empirical Methods in Natural Language Processing}, pp.\  8273--8288, Abu Dhabi, United Arab Emirates, December 2022. Association for Computational Linguistics.
\newblock \doi{10.18653/v1/2022.emnlp-main.566}.
\newblock URL \url{https://aclanthology.org/2022.emnlp-main.566/}.

\bibitem[Sun et~al.(2023)Sun, Fang, Wu, Wang, and Cao]{eva-clip}
Quan Sun, Yuxin Fang, Ledell Wu, Xinlong Wang, and Yue Cao.
\newblock Eva-clip: Improved training techniques for clip at scale, 2023.
\newblock URL \url{https://arxiv.org/abs/2303.15389}.

\bibitem[Team et~al.(2025)Team, Anil, Borgeaud, Alayrac, Yu, Soricut, Schalkwyk, Dai, Hauth, Millican, Silver, Johnson, Antonoglou, Schrittwieser, Glaese, Chen, Pitler, Lillicrap, Lazaridou, Firat, Molloy, Isard, Barham, Hennigan, Lee, Viola, Reynolds, Xu, Doherty, Collins, Meyer, Rutherford, Moreira, Ayoub, Goel, Krawczyk, Du, Chi, Cheng, Ni, Shah, Kane, Chan, Faruqui, Severyn, Lin, Li, Cheng, Ittycheriah, Mahdieh, Chen, Sun, Tran, Bagri, Lakshminarayanan, Liu, Orban, Güra, Zhou, Song, Boffy, Ganapathy, Zheng, Choe, Ágoston Weisz, Zhu, Lu, Gopal, Kahn, Kula, Pitman, Shah, Taropa, Merey, Baeuml, Chen, Shafey, Zhang, Sercinoglu, Tucker, Piqueras, Krikun, Barr, Savinov, Danihelka, Roelofs, White, Andreassen, von Glehn, Yagati, Kazemi, Gonzalez, Khalman, Sygnowski, Frechette, Smith, Culp, Proleev, Luan, Chen, Lottes, Schucher, Lebron, Rrustemi, Clay, Crone, Kocisky, Zhao, Perz, Yu, Howard, Bloniarz, Rae, Lu, Sifre, Maggioni, Alcober, Garrette, Barnes, Thakoor, Austin, Barth-Maron, Wong, Joshi, Chaabouni,
  Fatiha, Ahuja, Tomar, Senter, Chadwick, Kornakov, Attaluri, Iturrate, Liu, Li, Cogan, Chen, Jia, Gu, Zhang, Grimstad, Hartman, Garcia, Pillai, Devlin, Laskin, de~Las~Casas, Valter, Tao, Blanco, Badia, Reitter, Chen, Brennan, Rivera, Brin, Iqbal, Surita, Labanowski, Rao, Winkler, Parisotto, Gu, Olszewska, Addanki, Miech, Louis, Teplyashin, Brown, Catt, Balaguer, Xiang, Wang, Ashwood, Briukhov, Webson, Ganapathy, Sanghavi, Kannan, Chang, Stjerngren, Djolonga, Sun, Bapna, Aitchison, Pejman, Michalewski, Yu, Wang, Love, Ahn, Bloxwich, Han, Humphreys, Sellam, Bradbury, Godbole, Samangooei, Damoc, Kaskasoli, Arnold, Vasudevan, Agrawal, Riesa, Lepikhin, Tanburn, Srinivasan, Lim, Hodkinson, Shyam, Ferret, Hand, Garg, Paine, Li, Li, Giang, Neitz, Abbas, York, Reid, Cole, Chowdhery, Das, Rogozińska, Nikolaev, Sprechmann, Nado, Zilka, Prost, He, Monteiro, Mishra, Welty, Newlan, Jia, Allamanis, Hu, de~Liedekerke, Gilmer, Saroufim, Rijhwani, Hou, Shrivastava, Baddepudi, Goldin, Ozturel, Cassirer, Xu, Sohn, Sachan,
  Amplayo, Swanson, Petrova, Narayan, Guez, Brahma, Landon, Patel, Zhao, Villela, Wang, Jia, Rahtz, Giménez, Yeung, Keeling, Georgiev, Mincu, Wu, Haykal, Saputro, Vodrahalli, Qin, Cankara, Sharma, Fernando, Hawkins, Neyshabur, Kim, Hutter, Agrawal, Castro-Ros, van~den Driessche, Wang, Yang, yiin Chang, Komarek, McIlroy, Lučić, Zhang, Farhan, Sharman, Natsev, Michel, Bansal, Qiao, Cao, Shakeri, Butterfield, Chung, Rubenstein, Agrawal, Mensch, Soparkar, Lenc, Chung, Pope, Maggiore, Kay, Jhakra, Wang, Maynez, Phuong, Tobin, Tacchetti, Trebacz, Robinson, Katariya, Riedel, Bailey, Xiao, Ghelani, Aroyo, Slone, Houlsby, Xiong, Yang, Gribovskaya, Adler, Wirth, Lee, Li, Kagohara, Pavagadhi, Bridgers, Bortsova, Ghemawat, Ahmed, Liu, Powell, Bolina, Iinuma, Zablotskaia, Besley, Chung, Dozat, Comanescu, Si, Greer, Su, Polacek, Kaufman, Tokumine, Hu, Buchatskaya, Miao, Elhawaty, Siddhant, Tomasev, Xing, Greer, Miller, Ashraf, Roy, Zhang, Ma, Filos, Besta, Blevins, Klimenko, Yeh, Changpinyo, Mu, Chang, Pajarskas, Muir,
  Cohen, Lan, Haridasan, Marathe, Hansen, Douglas, Samuel, Wang, Austin, Lan, Jiang, Chiu, Lorenzo, Sjösund, Cevey, Gleicher, Avrahami, Boral, Srinivasan, Selo, May, Aisopos, Hussenot, Soares, Baumli, Chang, Recasens, Caine, Pritzel, Pavetic, Pardo, Gergely, Frye, Ramasesh, Horgan, Badola, Kassner, Roy, Dyer, Campos, Tomala, Tang, Badawy, White, Mustafa, Lang, Jindal, Vikram, Gong, Caelles, Hemsley, Thornton, Feng, Stokowiec, Zheng, Thacker, Çağlar Ünlü, Zhang, Saleh, Svensson, Bileschi, Patil, Anand, Ring, Tsihlas, Vezer, Selvi, Shevlane, Rodriguez, Kwiatkowski, Daruki, Rong, Dafoe, FitzGerald, Gu-Lemberg, Khan, Hendricks, Pellat, Feinberg, Cobon-Kerr, Sainath, Rauh, Hashemi, Ives, Hasson, Noland, Cao, Byrd, Hou, Wang, Sottiaux, Paganini, Lespiau, Moufarek, Hassan, Shivakumar, van Amersfoort, Mandhane, Joshi, Goyal, Tung, Brock, Sheahan, Misra, Li, Rakićević, Dehghani, Liu, Mittal, Oh, Noury, Sezener, Huot, Lamm, Cao, Chen, Mudgal, Stella, Brooks, Vasudevan, Liu, Chain, Melinkeri, Cohen, Wang,
  Seymore, Zubkov, Goel, Yue, Krishnakumaran, Albert, Hurley, Sano, Mohananey, Joughin, Filonov, Kępa, Eldawy, Lim, Rishi, Badiezadegan, Bos, Chang, Jain, Padmanabhan, Puttagunta, Krishna, Baker, Kalb, Bedapudi, Kurzrok, Lei, Yu, Litvin, Zhou, Wu, Sobell, Siciliano, Papir, Neale, Bragagnolo, Toor, Chen, Anklin, Wang, Feng, Gholami, Ling, Liu, Walter, Moghaddam, Kishore, Adamek, Mercado, Mallinson, Wandekar, Cagle, Ofek, Garrido, Lombriser, Mukha, Sun, Mohammad, Matak, Qian, Peswani, Janus, Yuan, Schelin, David, Garg, He, Duzhyi, Älgmyr, Lottaz, Li, Yadav, Xu, Chinien, Shivanna, Chuklin, Li, Spadine, Wolfe, Mohamed, Das, Dai, He, von Dincklage, Upadhyay, Maurya, Chi, Krause, Salama, Rabinovitch, M, Selvan, Dektiarev, Ghiasi, Guven, Gupta, Liu, Sharma, Shtacher, Paul, Akerlund, Aubet, Huang, Zhu, Zhu, Teixeira, Fritze, Bertolini, Marinescu, Bölle, Paulus, Gupta, Latkar, Chang, Sanders, Wilson, Wu, Tan, Thiet, Doshi, Lall, Mishra, Chen, Luong, Benjamin, Lee, Andrejczuk, Rabiej, Ranjan, Styrc, Yin, Simon,
  Harriott, Bansal, Robsky, Bacon, Greene, Mirylenka, Zhou, Sarvana, Goyal, Andermatt, Siegler, Horn, Israel, Pongetti, Chen, Selvatici, Silva, Wang, Tolins, Guu, Yogev, Cai, Agostini, Shah, Nguyen, Donnaile, Pereira, Friso, Stambler, Kurzrok, Kuang, Romanikhin, Geller, Yan, Jang, Lee, Fica, Malmi, Tan, Banica, Balle, Pham, Huang, Avram, Shi, Singh, Hidey, Ahuja, Saxena, Dooley, Potharaju, O'Neill, Gokulchandran, Foley, Zhao, Dusenberry, Liu, Mehta, Kotikalapudi, Safranek-Shrader, Goodman, Kessinger, Globen, Kolhar, Gorgolewski, Ibrahim, Song, Eichenbaum, Brovelli, Potluri, Lahoti, Baetu, Ghorbani, Chen, Crawford, Pal, Sridhar, Gurita, Mujika, Petrovski, Cedoz, Li, Chen, Santo, Goyal, Punjabi, Kappaganthu, Kwak, LV, Velury, Choudhury, Hall, Shah, Figueira, Thomas, Lu, Zhou, Kumar, Jurdi, Chikkerur, Ma, Yu, Kwak, Ähdel, Rajayogam, Choma, Liu, Barua, Ji, Park, Hellendoorn, Bailey, Bilal, Zhou, Khatir, Sutton, Rzadkowski, Macintosh, Vij, Shagin, Medina, Liang, Zhou, Shah, Bi, Dankovics, Banga, Lehmann,
  Bredesen, Lin, Hoffmann, Lai, Chung, Yang, Balani, Bražinskas, Sozanschi, Hayes, Alcalde, Makarov, Chen, Stella, Snijders, Mandl, Kärrman, Nowak, Wu, Dyck, Vaidyanathan, R, Mallet, Rudominer, Johnston, Mittal, Udathu, Christensen, Verma, Irving, Santucci, Elsayed, Davoodi, Georgiev, Tenney, Hua, Cideron, Leurent, Alnahlawi, Georgescu, Wei, Zheng, Scandinaro, Jiang, Snoek, Sundararajan, Wang, Ontiveros, Karo, Cole, Rajashekhar, Tumeh, Ben-David, Jain, Uesato, Datta, Bunyan, Wu, Zhang, Stanczyk, Zhang, Steiner, Naskar, Azzam, Johnson, Paszke, Chiu, Elias, Mohiuddin, Muhammad, Miao, Lee, Vieillard, Park, Zhang, Stanway, Garmon, Karmarkar, Dong, Lee, Kumar, Zhou, Evens, Isaac, Irving, Loper, Fink, Arkatkar, Chen, Shafran, Petrychenko, Chen, Jia, Levskaya, Zhu, Grabowski, Mao, Magni, Yao, Snaider, Casagrande, Palmer, Suganthan, Castaño, Giannoumis, Kim, Rybiński, Sreevatsa, Prendki, Soergel, Goedeckemeyer, Gierke, Jafari, Gaba, Wiesner, Wright, Wei, Vashisht, Kulizhskaya, Hoover, Le, Li, Iwuanyanwu, Liu,
  Ramirez, Khorlin, Cui, LIN, Wu, Aguilar, Pallo, Chakladar, Perng, Abellan, Zhang, Dasgupta, Kushman, Penchev, Repina, Wu, van~der Weide, Ponnapalli, Kaplan, Simsa, Li, Dousse, Yang, Piper, Ie, Pasumarthi, Lintz, Vijayakumar, Andor, Valenzuela, Lui, Paduraru, Peng, Lee, Zhang, Greene, Nguyen, Kurylowicz, Hardin, Dixon, Janzer, Choo, Feng, Zhang, Singhal, Du, McKinnon, Antropova, Bolukbasi, Keller, Reid, Finchelstein, Raad, Crocker, Hawkins, Dadashi, Gaffney, Franko, Bulanova, Leblond, Chung, Askham, Cobo, Xu, Fischer, Xu, Sorokin, Alberti, Lin, Evans, Dimitriev, Forbes, Banarse, Tung, Omernick, Bishop, Sterneck, Jain, Xia, Amid, Piccinno, Wang, Banzal, Mankowitz, Polozov, Krakovna, Brown, Bateni, Duan, Firoiu, Thotakuri, Natan, Geist, tan Girgin, Li, Ye, Roval, Tojo, Kwong, Lee-Thorp, Yew, Sinopalnikov, Ramos, Mellor, Sharma, Wu, Miller, Sonnerat, Vnukov, Greig, Beattie, Caveness, Bai, Eisenschlos, Korchemniy, Tsai, Jasarevic, Kong, Dao, Zheng, Liu, Yang, Zhu, Teh, Sanmiya, Gladchenko, Trdin, Toyama, Rosen,
  Tavakkol, Xue, Elkind, Woodman, Carpenter, Papamakarios, Kemp, Kafle, Grunina, Sinha, Talbert, Wu, Owusu-Afriyie, Du, Thornton, Pont-Tuset, Narayana, Li, Fatehi, Wieting, Ajmeri, Uria, Ko, Knight, Héliou, Niu, Gu, Pang, Li, Levine, Stolovich, Santamaria-Fernandez, Goenka, Yustalim, Strudel, Elqursh, Deck, Lee, Li, Levin, Hoffmann, Holtmann-Rice, Bachem, Arora, Koh, Yeganeh, Põder, Tariq, Sun, Ionita, Seyedhosseini, Tafti, Liu, Gulati, Liu, Ye, Chrzaszcz, Wang, Sethi, Li, Brown, Singh, Fan, Parisi, Stanton, Koverkathu, Choquette-Choo, Li, Lu, Ittycheriah, Shroff, Varadarajan, Bahargam, Willoughby, Gaddy, Desjardins, Cornero, Robenek, Mittal, Albrecht, Shenoy, Moiseev, Jacobsson, Ghaffarkhah, Rivière, Walton, Crepy, Parrish, Zhou, Farabet, Radebaugh, Srinivasan, van~der Salm, Fidjeland, Scellato, Latorre-Chimoto, Klimczak-Plucińska, Bridson, de~Cesare, Hudson, Mendolicchio, Walker, Morris, Mauger, Guseynov, Reid, Odoom, Loher, Cotruta, Yenugula, Grewe, Petrushkina, Duerig, Sanchez, Yadlowsky, Shen,
  Globerson, Webb, Dua, Li, Bhupatiraju, Hurt, Qureshi, Agarwal, Shani, Eyal, Khare, Belle, Wang, Tekur, Kale, Wei, Sang, Saeta, Liechty, Sun, Zhao, Lee, Nayak, Fritz, Vuyyuru, Aslanides, Vyas, Wicke, Ma, Eltyshev, Martin, Cate, Manyika, Amiri, Kim, Xiong, Kang, Luisier, Tripuraneni, Madras, Guo, Waters, Wang, Ainslie, Baldridge, Zhang, Pruthi, Bauer, Yang, Mansour, Gelman, Xu, Polovets, Liu, Cai, Chen, Sheng, Xue, Ozair, Angermueller, Li, Sinha, Wang, Wiesinger, Koukoumidis, Tian, Iyer, Gurumurthy, Goldenson, Shah, Blake, Yu, Urbanowicz, Palomaki, Fernando, Durden, Mehta, Momchev, Rahimtoroghi, Georgaki, Raul, Ruder, Redshaw, Lee, Zhou, Jalan, Li, Hechtman, Schuh, Nasr, Milan, Mikulik, Franco, Green, Nguyen, Kelley, Mahendru, Hu, Howland, Vargas, Hui, Bansal, Rao, Ghiya, Wang, Ye, Sarr, Preston, Elish, Li, Kaku, Gupta, Pasupat, Juan, Someswar, M., Chen, Amini, Fabrikant, Chu, Dong, Muthal, Buthpitiya, Jauhari, Hua, Khandelwal, Hitron, Ren, Rinaldi, Drath, Dabush, Jiang, Godhia, Sachs, Chen, Fan, Taitelbaum,
  Noga, Dai, Wang, Liang, Hamer, Ferng, Elkind, Atias, Lee, Listík, Carlen, van~de Kerkhof, Pikus, Zaher, Müller, Zykova, Stefanec, Gatsko, Hirnschall, Sethi, Xu, Ahuja, Tsai, Stefanoiu, Feng, Dhandhania, Katyal, Gupta, Parulekar, Pitta, Zhao, Bhatia, Bhavnani, Alhadlaq, Li, Danenberg, Tu, Pine, Filippova, Ghosh, Limonchik, Urala, Lanka, Clive, Sun, Li, Wu, Hongtongsak, Li, Thakkar, Omarov, Majmundar, Alverson, Kucharski, Patel, Jain, Zabelin, Pelagatti, Kohli, Kumar, Kim, Sankar, Shah, Ramachandruni, Zeng, Bariach, Weidinger, Vu, Andreev, He, Hui, Kashem, Subramanya, Hsiao, Hassabis, Kavukcuoglu, Sadovsky, Le, Strohman, Wu, Petrov, Dean, and Vinyals]{geminiteam2025geminifamilyhighlycapable}
Gemini Team, Rohan Anil, Sebastian Borgeaud, Jean-Baptiste Alayrac, Jiahui Yu, Radu Soricut, Johan Schalkwyk, Andrew~M. Dai, Anja Hauth, Katie Millican, David Silver, Melvin Johnson, Ioannis Antonoglou, Julian Schrittwieser, Amelia Glaese, Jilin Chen, Emily Pitler, Timothy Lillicrap, Angeliki Lazaridou, Orhan Firat, James Molloy, Michael Isard, Paul~R. Barham, Tom Hennigan, Benjamin Lee, Fabio Viola, Malcolm Reynolds, Yuanzhong Xu, Ryan Doherty, Eli Collins, Clemens Meyer, Eliza Rutherford, Erica Moreira, Kareem Ayoub, Megha Goel, Jack Krawczyk, Cosmo Du, Ed~Chi, Heng-Tze Cheng, Eric Ni, Purvi Shah, Patrick Kane, Betty Chan, Manaal Faruqui, Aliaksei Severyn, Hanzhao Lin, YaGuang Li, Yong Cheng, Abe Ittycheriah, Mahdis Mahdieh, Mia Chen, Pei Sun, Dustin Tran, Sumit Bagri, Balaji Lakshminarayanan, Jeremiah Liu, Andras Orban, Fabian Güra, Hao Zhou, Xinying Song, Aurelien Boffy, Harish Ganapathy, Steven Zheng, HyunJeong Choe, Ágoston Weisz, Tao Zhu, Yifeng Lu, Siddharth Gopal, Jarrod Kahn, Maciej Kula, Jeff
  Pitman, Rushin Shah, Emanuel Taropa, Majd~Al Merey, Martin Baeuml, Zhifeng Chen, Laurent~El Shafey, Yujing Zhang, Olcan Sercinoglu, George Tucker, Enrique Piqueras, Maxim Krikun, Iain Barr, Nikolay Savinov, Ivo Danihelka, Becca Roelofs, Anaïs White, Anders Andreassen, Tamara von Glehn, Lakshman Yagati, Mehran Kazemi, Lucas Gonzalez, Misha Khalman, Jakub Sygnowski, Alexandre Frechette, Charlotte Smith, Laura Culp, Lev Proleev, Yi~Luan, Xi~Chen, James Lottes, Nathan Schucher, Federico Lebron, Alban Rrustemi, Natalie Clay, Phil Crone, Tomas Kocisky, Jeffrey Zhao, Bartek Perz, Dian Yu, Heidi Howard, Adam Bloniarz, Jack~W. Rae, Han Lu, Laurent Sifre, Marcello Maggioni, Fred Alcober, Dan Garrette, Megan Barnes, Shantanu Thakoor, Jacob Austin, Gabriel Barth-Maron, William Wong, Rishabh Joshi, Rahma Chaabouni, Deeni Fatiha, Arun Ahuja, Gaurav~Singh Tomar, Evan Senter, Martin Chadwick, Ilya Kornakov, Nithya Attaluri, Iñaki Iturrate, Ruibo Liu, Yunxuan Li, Sarah Cogan, Jeremy Chen, Chao Jia, Chenjie Gu, Qiao Zhang,
  Jordan Grimstad, Ale~Jakse Hartman, Xavier Garcia, Thanumalayan~Sankaranarayana Pillai, Jacob Devlin, Michael Laskin, Diego de~Las~Casas, Dasha Valter, Connie Tao, Lorenzo Blanco, Adrià~Puigdomènech Badia, David Reitter, Mianna Chen, Jenny Brennan, Clara Rivera, Sergey Brin, Shariq Iqbal, Gabriela Surita, Jane Labanowski, Abhi Rao, Stephanie Winkler, Emilio Parisotto, Yiming Gu, Kate Olszewska, Ravi Addanki, Antoine Miech, Annie Louis, Denis Teplyashin, Geoff Brown, Elliot Catt, Jan Balaguer, Jackie Xiang, Pidong Wang, Zoe Ashwood, Anton Briukhov, Albert Webson, Sanjay Ganapathy, Smit Sanghavi, Ajay Kannan, Ming-Wei Chang, Axel Stjerngren, Josip Djolonga, Yuting Sun, Ankur Bapna, Matthew Aitchison, Pedram Pejman, Henryk Michalewski, Tianhe Yu, Cindy Wang, Juliette Love, Junwhan Ahn, Dawn Bloxwich, Kehang Han, Peter Humphreys, Thibault Sellam, James Bradbury, Varun Godbole, Sina Samangooei, Bogdan Damoc, Alex Kaskasoli, Sébastien M.~R. Arnold, Vijay Vasudevan, Shubham Agrawal, Jason Riesa, Dmitry
  Lepikhin, Richard Tanburn, Srivatsan Srinivasan, Hyeontaek Lim, Sarah Hodkinson, Pranav Shyam, Johan Ferret, Steven Hand, Ankush Garg, Tom~Le Paine, Jian Li, Yujia Li, Minh Giang, Alexander Neitz, Zaheer Abbas, Sarah York, Machel Reid, Elizabeth Cole, Aakanksha Chowdhery, Dipanjan Das, Dominika Rogozińska, Vitaliy Nikolaev, Pablo Sprechmann, Zachary Nado, Lukas Zilka, Flavien Prost, Luheng He, Marianne Monteiro, Gaurav Mishra, Chris Welty, Josh Newlan, Dawei Jia, Miltiadis Allamanis, Clara~Huiyi Hu, Raoul de~Liedekerke, Justin Gilmer, Carl Saroufim, Shruti Rijhwani, Shaobo Hou, Disha Shrivastava, Anirudh Baddepudi, Alex Goldin, Adnan Ozturel, Albin Cassirer, Yunhan Xu, Daniel Sohn, Devendra Sachan, Reinald~Kim Amplayo, Craig Swanson, Dessie Petrova, Shashi Narayan, Arthur Guez, Siddhartha Brahma, Jessica Landon, Miteyan Patel, Ruizhe Zhao, Kevin Villela, Luyu Wang, Wenhao Jia, Matthew Rahtz, Mai Giménez, Legg Yeung, James Keeling, Petko Georgiev, Diana Mincu, Boxi Wu, Salem Haykal, Rachel Saputro, Kiran
  Vodrahalli, James Qin, Zeynep Cankara, Abhanshu Sharma, Nick Fernando, Will Hawkins, Behnam Neyshabur, Solomon Kim, Adrian Hutter, Priyanka Agrawal, Alex Castro-Ros, George van~den Driessche, Tao Wang, Fan Yang, Shuo yiin Chang, Paul Komarek, Ross McIlroy, Mario Lučić, Guodong Zhang, Wael Farhan, Michael Sharman, Paul Natsev, Paul Michel, Yamini Bansal, Siyuan Qiao, Kris Cao, Siamak Shakeri, Christina Butterfield, Justin Chung, Paul~Kishan Rubenstein, Shivani Agrawal, Arthur Mensch, Kedar Soparkar, Karel Lenc, Timothy Chung, Aedan Pope, Loren Maggiore, Jackie Kay, Priya Jhakra, Shibo Wang, Joshua Maynez, Mary Phuong, Taylor Tobin, Andrea Tacchetti, Maja Trebacz, Kevin Robinson, Yash Katariya, Sebastian Riedel, Paige Bailey, Kefan Xiao, Nimesh Ghelani, Lora Aroyo, Ambrose Slone, Neil Houlsby, Xuehan Xiong, Zhen Yang, Elena Gribovskaya, Jonas Adler, Mateo Wirth, Lisa Lee, Music Li, Thais Kagohara, Jay Pavagadhi, Sophie Bridgers, Anna Bortsova, Sanjay Ghemawat, Zafarali Ahmed, Tianqi Liu, Richard Powell,
  Vijay Bolina, Mariko Iinuma, Polina Zablotskaia, James Besley, Da-Woon Chung, Timothy Dozat, Ramona Comanescu, Xiance Si, Jeremy Greer, Guolong Su, Martin Polacek, Raphaël~Lopez Kaufman, Simon Tokumine, Hexiang Hu, Elena Buchatskaya, Yingjie Miao, Mohamed Elhawaty, Aditya Siddhant, Nenad Tomasev, Jinwei Xing, Christina Greer, Helen Miller, Shereen Ashraf, Aurko Roy, Zizhao Zhang, Ada Ma, Angelos Filos, Milos Besta, Rory Blevins, Ted Klimenko, Chih-Kuan Yeh, Soravit Changpinyo, Jiaqi Mu, Oscar Chang, Mantas Pajarskas, Carrie Muir, Vered Cohen, Charline~Le Lan, Krishna Haridasan, Amit Marathe, Steven Hansen, Sholto Douglas, Rajkumar Samuel, Mingqiu Wang, Sophia Austin, Chang Lan, Jiepu Jiang, Justin Chiu, Jaime~Alonso Lorenzo, Lars~Lowe Sjösund, Sébastien Cevey, Zach Gleicher, Thi Avrahami, Anudhyan Boral, Hansa Srinivasan, Vittorio Selo, Rhys May, Konstantinos Aisopos, Léonard Hussenot, Livio~Baldini Soares, Kate Baumli, Michael~B. Chang, Adrià Recasens, Ben Caine, Alexander Pritzel, Filip Pavetic,
  Fabio Pardo, Anita Gergely, Justin Frye, Vinay Ramasesh, Dan Horgan, Kartikeya Badola, Nora Kassner, Subhrajit Roy, Ethan Dyer, Víctor~Campos Campos, Alex Tomala, Yunhao Tang, Dalia~El Badawy, Elspeth White, Basil Mustafa, Oran Lang, Abhishek Jindal, Sharad Vikram, Zhitao Gong, Sergi Caelles, Ross Hemsley, Gregory Thornton, Fangxiaoyu Feng, Wojciech Stokowiec, Ce~Zheng, Phoebe Thacker, Çağlar Ünlü, Zhishuai Zhang, Mohammad Saleh, James Svensson, Max Bileschi, Piyush Patil, Ankesh Anand, Roman Ring, Katerina Tsihlas, Arpi Vezer, Marco Selvi, Toby Shevlane, Mikel Rodriguez, Tom Kwiatkowski, Samira Daruki, Keran Rong, Allan Dafoe, Nicholas FitzGerald, Keren Gu-Lemberg, Mina Khan, Lisa~Anne Hendricks, Marie Pellat, Vladimir Feinberg, James Cobon-Kerr, Tara Sainath, Maribeth Rauh, Sayed~Hadi Hashemi, Richard Ives, Yana Hasson, Eric Noland, Yuan Cao, Nathan Byrd, Le~Hou, Qingze Wang, Thibault Sottiaux, Michela Paganini, Jean-Baptiste Lespiau, Alexandre Moufarek, Samer Hassan, Kaushik Shivakumar, Joost van
  Amersfoort, Amol Mandhane, Pratik Joshi, Anirudh Goyal, Matthew Tung, Andrew Brock, Hannah Sheahan, Vedant Misra, Cheng Li, Nemanja Rakićević, Mostafa Dehghani, Fangyu Liu, Sid Mittal, Junhyuk Oh, Seb Noury, Eren Sezener, Fantine Huot, Matthew Lamm, Nicola~De Cao, Charlie Chen, Sidharth Mudgal, Romina Stella, Kevin Brooks, Gautam Vasudevan, Chenxi Liu, Mainak Chain, Nivedita Melinkeri, Aaron Cohen, Venus Wang, Kristie Seymore, Sergey Zubkov, Rahul Goel, Summer Yue, Sai Krishnakumaran, Brian Albert, Nate Hurley, Motoki Sano, Anhad Mohananey, Jonah Joughin, Egor Filonov, Tomasz Kępa, Yomna Eldawy, Jiawern Lim, Rahul Rishi, Shirin Badiezadegan, Taylor Bos, Jerry Chang, Sanil Jain, Sri Gayatri~Sundara Padmanabhan, Subha Puttagunta, Kalpesh Krishna, Leslie Baker, Norbert Kalb, Vamsi Bedapudi, Adam Kurzrok, Shuntong Lei, Anthony Yu, Oren Litvin, Xiang Zhou, Zhichun Wu, Sam Sobell, Andrea Siciliano, Alan Papir, Robby Neale, Jonas Bragagnolo, Tej Toor, Tina Chen, Valentin Anklin, Feiran Wang, Richie Feng, Milad
  Gholami, Kevin Ling, Lijuan Liu, Jules Walter, Hamid Moghaddam, Arun Kishore, Jakub Adamek, Tyler Mercado, Jonathan Mallinson, Siddhinita Wandekar, Stephen Cagle, Eran Ofek, Guillermo Garrido, Clemens Lombriser, Maksim Mukha, Botu Sun, Hafeezul~Rahman Mohammad, Josip Matak, Yadi Qian, Vikas Peswani, Pawel Janus, Quan Yuan, Leif Schelin, Oana David, Ankur Garg, Yifan He, Oleksii Duzhyi, Anton Älgmyr, Timothée Lottaz, Qi~Li, Vikas Yadav, Luyao Xu, Alex Chinien, Rakesh Shivanna, Aleksandr Chuklin, Josie Li, Carrie Spadine, Travis Wolfe, Kareem Mohamed, Subhabrata Das, Zihang Dai, Kyle He, Daniel von Dincklage, Shyam Upadhyay, Akanksha Maurya, Luyan Chi, Sebastian Krause, Khalid Salama, Pam~G Rabinovitch, Pavan Kumar~Reddy M, Aarush Selvan, Mikhail Dektiarev, Golnaz Ghiasi, Erdem Guven, Himanshu Gupta, Boyi Liu, Deepak Sharma, Idan~Heimlich Shtacher, Shachi Paul, Oscar Akerlund, François-Xavier Aubet, Terry Huang, Chen Zhu, Eric Zhu, Elico Teixeira, Matthew Fritze, Francesco Bertolini, Liana-Eleonora
  Marinescu, Martin Bölle, Dominik Paulus, Khyatti Gupta, Tejasi Latkar, Max Chang, Jason Sanders, Roopa Wilson, Xuewei Wu, Yi-Xuan Tan, Lam~Nguyen Thiet, Tulsee Doshi, Sid Lall, Swaroop Mishra, Wanming Chen, Thang Luong, Seth Benjamin, Jasmine Lee, Ewa Andrejczuk, Dominik Rabiej, Vipul Ranjan, Krzysztof Styrc, Pengcheng Yin, Jon Simon, Malcolm~Rose Harriott, Mudit Bansal, Alexei Robsky, Geoff Bacon, David Greene, Daniil Mirylenka, Chen Zhou, Obaid Sarvana, Abhimanyu Goyal, Samuel Andermatt, Patrick Siegler, Ben Horn, Assaf Israel, Francesco Pongetti, Chih-Wei~"Louis" Chen, Marco Selvatici, Pedro Silva, Kathie Wang, Jackson Tolins, Kelvin Guu, Roey Yogev, Xiaochen Cai, Alessandro Agostini, Maulik Shah, Hung Nguyen, Noah~Ó Donnaile, Sébastien Pereira, Linda Friso, Adam Stambler, Adam Kurzrok, Chenkai Kuang, Yan Romanikhin, Mark Geller, ZJ~Yan, Kane Jang, Cheng-Chun Lee, Wojciech Fica, Eric Malmi, Qijun Tan, Dan Banica, Daniel Balle, Ryan Pham, Yanping Huang, Diana Avram, Hongzhi Shi, Jasjot Singh, Chris
  Hidey, Niharika Ahuja, Pranab Saxena, Dan Dooley, Srividya~Pranavi Potharaju, Eileen O'Neill, Anand Gokulchandran, Ryan Foley, Kai Zhao, Mike Dusenberry, Yuan Liu, Pulkit Mehta, Ragha Kotikalapudi, Chalence Safranek-Shrader, Andrew Goodman, Joshua Kessinger, Eran Globen, Prateek Kolhar, Chris Gorgolewski, Ali Ibrahim, Yang Song, Ali Eichenbaum, Thomas Brovelli, Sahitya Potluri, Preethi Lahoti, Cip Baetu, Ali Ghorbani, Charles Chen, Andy Crawford, Shalini Pal, Mukund Sridhar, Petru Gurita, Asier Mujika, Igor Petrovski, Pierre-Louis Cedoz, Chenmei Li, Shiyuan Chen, Niccolò~Dal Santo, Siddharth Goyal, Jitesh Punjabi, Karthik Kappaganthu, Chester Kwak, Pallavi LV, Sarmishta Velury, Himadri Choudhury, Jamie Hall, Premal Shah, Ricardo Figueira, Matt Thomas, Minjie Lu, Ting Zhou, Chintu Kumar, Thomas Jurdi, Sharat Chikkerur, Yenai Ma, Adams Yu, Soo Kwak, Victor Ähdel, Sujeevan Rajayogam, Travis Choma, Fei Liu, Aditya Barua, Colin Ji, Ji~Ho Park, Vincent Hellendoorn, Alex Bailey, Taylan Bilal, Huanjie Zhou,
  Mehrdad Khatir, Charles Sutton, Wojciech Rzadkowski, Fiona Macintosh, Roopali Vij, Konstantin Shagin, Paul Medina, Chen Liang, Jinjing Zhou, Pararth Shah, Yingying Bi, Attila Dankovics, Shipra Banga, Sabine Lehmann, Marissa Bredesen, Zifan Lin, John~Eric Hoffmann, Jonathan Lai, Raynald Chung, Kai Yang, Nihal Balani, Arthur Bražinskas, Andrei Sozanschi, Matthew Hayes, Héctor~Fernández Alcalde, Peter Makarov, Will Chen, Antonio Stella, Liselotte Snijders, Michael Mandl, Ante Kärrman, Paweł Nowak, Xinyi Wu, Alex Dyck, Krishnan Vaidyanathan, Raghavender R, Jessica Mallet, Mitch Rudominer, Eric Johnston, Sushil Mittal, Akhil Udathu, Janara Christensen, Vishal Verma, Zach Irving, Andreas Santucci, Gamaleldin Elsayed, Elnaz Davoodi, Marin Georgiev, Ian Tenney, Nan Hua, Geoffrey Cideron, Edouard Leurent, Mahmoud Alnahlawi, Ionut Georgescu, Nan Wei, Ivy Zheng, Dylan Scandinaro, Heinrich Jiang, Jasper Snoek, Mukund Sundararajan, Xuezhi Wang, Zack Ontiveros, Itay Karo, Jeremy Cole, Vinu Rajashekhar, Lara Tumeh,
  Eyal Ben-David, Rishub Jain, Jonathan Uesato, Romina Datta, Oskar Bunyan, Shimu Wu, John Zhang, Piotr Stanczyk, Ye~Zhang, David Steiner, Subhajit Naskar, Michael Azzam, Matthew Johnson, Adam Paszke, Chung-Cheng Chiu, Jaume~Sanchez Elias, Afroz Mohiuddin, Faizan Muhammad, Jin Miao, Andrew Lee, Nino Vieillard, Jane Park, Jiageng Zhang, Jeff Stanway, Drew Garmon, Abhijit Karmarkar, Zhe Dong, Jong Lee, Aviral Kumar, Luowei Zhou, Jonathan Evens, William Isaac, Geoffrey Irving, Edward Loper, Michael Fink, Isha Arkatkar, Nanxin Chen, Izhak Shafran, Ivan Petrychenko, Zhe Chen, Johnson Jia, Anselm Levskaya, Zhenkai Zhu, Peter Grabowski, Yu~Mao, Alberto Magni, Kaisheng Yao, Javier Snaider, Norman Casagrande, Evan Palmer, Paul Suganthan, Alfonso Castaño, Irene Giannoumis, Wooyeol Kim, Mikołaj Rybiński, Ashwin Sreevatsa, Jennifer Prendki, David Soergel, Adrian Goedeckemeyer, Willi Gierke, Mohsen Jafari, Meenu Gaba, Jeremy Wiesner, Diana~Gage Wright, Yawen Wei, Harsha Vashisht, Yana Kulizhskaya, Jay Hoover, Maigo Le,
  Lu~Li, Chimezie Iwuanyanwu, Lu~Liu, Kevin Ramirez, Andrey Khorlin, Albert Cui, Tian LIN, Marcus Wu, Ricardo Aguilar, Keith Pallo, Abhishek Chakladar, Ginger Perng, Elena~Allica Abellan, Mingyang Zhang, Ishita Dasgupta, Nate Kushman, Ivo Penchev, Alena Repina, Xihui Wu, Tom van~der Weide, Priya Ponnapalli, Caroline Kaplan, Jiri Simsa, Shuangfeng Li, Olivier Dousse, Fan Yang, Jeff Piper, Nathan Ie, Rama Pasumarthi, Nathan Lintz, Anitha Vijayakumar, Daniel Andor, Pedro Valenzuela, Minnie Lui, Cosmin Paduraru, Daiyi Peng, Katherine Lee, Shuyuan Zhang, Somer Greene, Duc~Dung Nguyen, Paula Kurylowicz, Cassidy Hardin, Lucas Dixon, Lili Janzer, Kiam Choo, Ziqiang Feng, Biao Zhang, Achintya Singhal, Dayou Du, Dan McKinnon, Natasha Antropova, Tolga Bolukbasi, Orgad Keller, David Reid, Daniel Finchelstein, Maria~Abi Raad, Remi Crocker, Peter Hawkins, Robert Dadashi, Colin Gaffney, Ken Franko, Anna Bulanova, Rémi Leblond, Shirley Chung, Harry Askham, Luis~C. Cobo, Kelvin Xu, Felix Fischer, Jun Xu, Christina Sorokin,
  Chris Alberti, Chu-Cheng Lin, Colin Evans, Alek Dimitriev, Hannah Forbes, Dylan Banarse, Zora Tung, Mark Omernick, Colton Bishop, Rachel Sterneck, Rohan Jain, Jiawei Xia, Ehsan Amid, Francesco Piccinno, Xingyu Wang, Praseem Banzal, Daniel~J. Mankowitz, Alex Polozov, Victoria Krakovna, Sasha Brown, MohammadHossein Bateni, Dennis Duan, Vlad Firoiu, Meghana Thotakuri, Tom Natan, Matthieu Geist, Ser tan Girgin, Hui Li, Jiayu Ye, Ofir Roval, Reiko Tojo, Michael Kwong, James Lee-Thorp, Christopher Yew, Danila Sinopalnikov, Sabela Ramos, John Mellor, Abhishek Sharma, Kathy Wu, David Miller, Nicolas Sonnerat, Denis Vnukov, Rory Greig, Jennifer Beattie, Emily Caveness, Libin Bai, Julian Eisenschlos, Alex Korchemniy, Tomy Tsai, Mimi Jasarevic, Weize Kong, Phuong Dao, Zeyu Zheng, Frederick Liu, Fan Yang, Rui Zhu, Tian~Huey Teh, Jason Sanmiya, Evgeny Gladchenko, Nejc Trdin, Daniel Toyama, Evan Rosen, Sasan Tavakkol, Linting Xue, Chen Elkind, Oliver Woodman, John Carpenter, George Papamakarios, Rupert Kemp, Sushant
  Kafle, Tanya Grunina, Rishika Sinha, Alice Talbert, Diane Wu, Denese Owusu-Afriyie, Cosmo Du, Chloe Thornton, Jordi Pont-Tuset, Pradyumna Narayana, Jing Li, Saaber Fatehi, John Wieting, Omar Ajmeri, Benigno Uria, Yeongil Ko, Laura Knight, Amélie Héliou, Ning Niu, Shane Gu, Chenxi Pang, Yeqing Li, Nir Levine, Ariel Stolovich, Rebeca Santamaria-Fernandez, Sonam Goenka, Wenny Yustalim, Robin Strudel, Ali Elqursh, Charlie Deck, Hyo Lee, Zonglin Li, Kyle Levin, Raphael Hoffmann, Dan Holtmann-Rice, Olivier Bachem, Sho Arora, Christy Koh, Soheil~Hassas Yeganeh, Siim Põder, Mukarram Tariq, Yanhua Sun, Lucian Ionita, Mojtaba Seyedhosseini, Pouya Tafti, Zhiyu Liu, Anmol Gulati, Jasmine Liu, Xinyu Ye, Bart Chrzaszcz, Lily Wang, Nikhil Sethi, Tianrun Li, Ben Brown, Shreya Singh, Wei Fan, Aaron Parisi, Joe Stanton, Vinod Koverkathu, Christopher~A. Choquette-Choo, Yunjie Li, TJ~Lu, Abe Ittycheriah, Prakash Shroff, Mani Varadarajan, Sanaz Bahargam, Rob Willoughby, David Gaddy, Guillaume Desjardins, Marco Cornero, Brona
  Robenek, Bhavishya Mittal, Ben Albrecht, Ashish Shenoy, Fedor Moiseev, Henrik Jacobsson, Alireza Ghaffarkhah, Morgane Rivière, Alanna Walton, Clément Crepy, Alicia Parrish, Zongwei Zhou, Clement Farabet, Carey Radebaugh, Praveen Srinivasan, Claudia van~der Salm, Andreas Fidjeland, Salvatore Scellato, Eri Latorre-Chimoto, Hanna Klimczak-Plucińska, David Bridson, Dario de~Cesare, Tom Hudson, Piermaria Mendolicchio, Lexi Walker, Alex Morris, Matthew Mauger, Alexey Guseynov, Alison Reid, Seth Odoom, Lucia Loher, Victor Cotruta, Madhavi Yenugula, Dominik Grewe, Anastasia Petrushkina, Tom Duerig, Antonio Sanchez, Steve Yadlowsky, Amy Shen, Amir Globerson, Lynette Webb, Sahil Dua, Dong Li, Surya Bhupatiraju, Dan Hurt, Haroon Qureshi, Ananth Agarwal, Tomer Shani, Matan Eyal, Anuj Khare, Shreyas~Rammohan Belle, Lei Wang, Chetan Tekur, Mihir~Sanjay Kale, Jinliang Wei, Ruoxin Sang, Brennan Saeta, Tyler Liechty, Yi~Sun, Yao Zhao, Stephan Lee, Pandu Nayak, Doug Fritz, Manish~Reddy Vuyyuru, John Aslanides, Nidhi Vyas,
  Martin Wicke, Xiao Ma, Evgenii Eltyshev, Nina Martin, Hardie Cate, James Manyika, Keyvan Amiri, Yelin Kim, Xi~Xiong, Kai Kang, Florian Luisier, Nilesh Tripuraneni, David Madras, Mandy Guo, Austin Waters, Oliver Wang, Joshua Ainslie, Jason Baldridge, Han Zhang, Garima Pruthi, Jakob Bauer, Feng Yang, Riham Mansour, Jason Gelman, Yang Xu, George Polovets, Ji~Liu, Honglong Cai, Warren Chen, XiangHai Sheng, Emily Xue, Sherjil Ozair, Christof Angermueller, Xiaowei Li, Anoop Sinha, Weiren Wang, Julia Wiesinger, Emmanouil Koukoumidis, Yuan Tian, Anand Iyer, Madhu Gurumurthy, Mark Goldenson, Parashar Shah, MK~Blake, Hongkun Yu, Anthony Urbanowicz, Jennimaria Palomaki, Chrisantha Fernando, Ken Durden, Harsh Mehta, Nikola Momchev, Elahe Rahimtoroghi, Maria Georgaki, Amit Raul, Sebastian Ruder, Morgan Redshaw, Jinhyuk Lee, Denny Zhou, Komal Jalan, Dinghua Li, Blake Hechtman, Parker Schuh, Milad Nasr, Kieran Milan, Vladimir Mikulik, Juliana Franco, Tim Green, Nam Nguyen, Joe Kelley, Aroma Mahendru, Andrea Hu, Joshua
  Howland, Ben Vargas, Jeffrey Hui, Kshitij Bansal, Vikram Rao, Rakesh Ghiya, Emma Wang, Ke~Ye, Jean~Michel Sarr, Melanie~Moranski Preston, Madeleine Elish, Steve Li, Aakash Kaku, Jigar Gupta, Ice Pasupat, Da-Cheng Juan, Milan Someswar, Tejvi M., Xinyun Chen, Aida Amini, Alex Fabrikant, Eric Chu, Xuanyi Dong, Amruta Muthal, Senaka Buthpitiya, Sarthak Jauhari, Nan Hua, Urvashi Khandelwal, Ayal Hitron, Jie Ren, Larissa Rinaldi, Shahar Drath, Avigail Dabush, Nan-Jiang Jiang, Harshal Godhia, Uli Sachs, Anthony Chen, Yicheng Fan, Hagai Taitelbaum, Hila Noga, Zhuyun Dai, James Wang, Chen Liang, Jenny Hamer, Chun-Sung Ferng, Chenel Elkind, Aviel Atias, Paulina Lee, Vít Listík, Mathias Carlen, Jan van~de Kerkhof, Marcin Pikus, Krunoslav Zaher, Paul Müller, Sasha Zykova, Richard Stefanec, Vitaly Gatsko, Christoph Hirnschall, Ashwin Sethi, Xingyu~Federico Xu, Chetan Ahuja, Beth Tsai, Anca Stefanoiu, Bo~Feng, Keshav Dhandhania, Manish Katyal, Akshay Gupta, Atharva Parulekar, Divya Pitta, Jing Zhao, Vivaan Bhatia,
  Yashodha Bhavnani, Omar Alhadlaq, Xiaolin Li, Peter Danenberg, Dennis Tu, Alex Pine, Vera Filippova, Abhipso Ghosh, Ben Limonchik, Bhargava Urala, Chaitanya~Krishna Lanka, Derik Clive, Yi~Sun, Edward Li, Hao Wu, Kevin Hongtongsak, Ianna Li, Kalind Thakkar, Kuanysh Omarov, Kushal Majmundar, Michael Alverson, Michael Kucharski, Mohak Patel, Mudit Jain, Maksim Zabelin, Paolo Pelagatti, Rohan Kohli, Saurabh Kumar, Joseph Kim, Swetha Sankar, Vineet Shah, Lakshmi Ramachandruni, Xiangkai Zeng, Ben Bariach, Laura Weidinger, Tu~Vu, Alek Andreev, Antoine He, Kevin Hui, Sheleem Kashem, Amar Subramanya, Sissie Hsiao, Demis Hassabis, Koray Kavukcuoglu, Adam Sadovsky, Quoc Le, Trevor Strohman, Yonghui Wu, Slav Petrov, Jeffrey Dean, and Oriol Vinyals.
\newblock Gemini: A family of highly capable multimodal models, 2025.
\newblock URL \url{https://arxiv.org/abs/2312.11805}.

\bibitem[Wang \& Liu(2021)Wang and Liu]{wang2021understandingbehaviourcontrastiveloss}
Feng Wang and Huaping Liu.
\newblock Understanding the behaviour of contrastive loss, 2021.
\newblock URL \url{https://arxiv.org/abs/2012.09740}.

\bibitem[Wang et~al.(2024{\natexlab{a}})Wang, Yang, Huang, Yang, Majumder, and Wei]{improvedtextembeddings}
Liang Wang, Nan Yang, Xiaolong Huang, Linjun Yang, Rangan Majumder, and Furu Wei.
\newblock Improving text embeddings with large language models, 2024{\natexlab{a}}.
\newblock URL \url{https://arxiv.org/abs/2401.00368}.

\bibitem[Wang et~al.(2024{\natexlab{b}})Wang, Bai, Tan, Wang, Fan, Bai, Chen, Liu, Wang, Ge, Fan, Dang, Du, Ren, Men, Liu, Zhou, Zhou, and Lin]{qwenvl}
Peng Wang, Shuai Bai, Sinan Tan, Shijie Wang, Zhihao Fan, Jinze Bai, Keqin Chen, Xuejing Liu, Jialin Wang, Wenbin Ge, Yang Fan, Kai Dang, Mengfei Du, Xuancheng Ren, Rui Men, Dayiheng Liu, Chang Zhou, Jingren Zhou, and Junyang Lin.
\newblock Qwen2-vl: Enhancing vision-language model's perception of the world at any resolution, 2024{\natexlab{b}}.
\newblock URL \url{https://arxiv.org/abs/2409.12191}.

\bibitem[Wei et~al.(2023)Wei, Chen, Chen, Hu, Zhang, Fu, Ritter, and Chen]{uniir}
Cong Wei, Yang Chen, Haonan Chen, Hexiang Hu, Ge~Zhang, Jie Fu, Alan Ritter, and Wenhu Chen.
\newblock Uniir: Training and benchmarking universal multimodal information retrievers, 2023.
\newblock URL \url{https://arxiv.org/abs/2311.17136}.

\bibitem[Wijmans et~al.(2024)Wijmans, Huval, Hertzberg, Koltun, and Krähenbühl]{ce_memory}
Erik Wijmans, Brody Huval, Alexander Hertzberg, Vladlen Koltun, and Philipp Krähenbühl.
\newblock Cut your losses in large-vocabulary language models, 2024.
\newblock URL \url{https://arxiv.org/abs/2411.09009}.

\bibitem[Xiao et~al.(2010)Xiao, Hays, Ehinger, Oliva, and Torralba]{sun397}
Jianxiong Xiao, James Hays, Krista~A. Ehinger, Aude Oliva, and Antonio Torralba.
\newblock Sun database: Large-scale scene recognition from abbey to zoo.
\newblock In \emph{2010 IEEE Computer Society Conference on Computer Vision and Pattern Recognition}, pp.\  3485--3492, 2010.
\newblock \doi{10.1109/CVPR.2010.5539970}.

\bibitem[Yao et~al.(2021)Yao, Huang, Hou, Lu, Niu, Xu, Liang, Li, Jiang, and Xu]{filip}
Lewei Yao, Runhui Huang, Lu~Hou, Guansong Lu, Minzhe Niu, Hang Xu, Xiaodan Liang, Zhenguo Li, Xin Jiang, and Chunjing Xu.
\newblock Filip: Fine-grained interactive language-image pre-training, 2021.
\newblock URL \url{https://arxiv.org/abs/2111.07783}.

\bibitem[Yu et~al.(2022)Yu, Wang, Vasudevan, Yeung, Seyedhosseini, and Wu]{coca}
Jiahui Yu, Zirui Wang, Vijay Vasudevan, Legg Yeung, Mojtaba Seyedhosseini, and Yonghui Wu.
\newblock Coca: Contrastive captioners are image-text foundation models, 2022.
\newblock URL \url{https://arxiv.org/abs/2205.01917}.

\bibitem[Yue et~al.(2024)Yue, Ni, Zhang, Zheng, Liu, Zhang, Stevens, Jiang, Ren, Sun, Wei, Yu, Yuan, Sun, Yin, Zheng, Yang, Liu, Huang, Sun, Su, and Chen]{mmmu}
Xiang Yue, Yuansheng Ni, Kai Zhang, Tianyu Zheng, Ruoqi Liu, Ge~Zhang, Samuel Stevens, Dongfu Jiang, Weiming Ren, Yuxuan Sun, Cong Wei, Botao Yu, Ruibin Yuan, Renliang Sun, Ming Yin, Boyuan Zheng, Zhenzhu Yang, Yibo Liu, Wenhao Huang, Huan Sun, Yu~Su, and Wenhu Chen.
\newblock Mmmu: A massive multi-discipline multimodal understanding and reasoning benchmark for expert agi, 2024.
\newblock URL \url{https://arxiv.org/abs/2311.16502}.

\bibitem[Zhang et~al.(2024)Zhang, Luan, Hu, Lee, Qiao, Chen, Su, and Chang]{magiclens}
Kai Zhang, Yi~Luan, Hexiang Hu, Kenton Lee, Siyuan Qiao, Wenhu Chen, Yu~Su, and Ming-Wei Chang.
\newblock Magiclens: Self-supervised image retrieval with open-ended instructions.
\newblock In \emph{The Forty-first International Conference on Machine Learning (ICML)}, pp.\  to appear, 2024.

\bibitem[Zhang et~al.(2025)Zhang, Zhang, Xie, Li, Dai, Long, Xie, Zhang, Li, and Zhang]{gme}
Xin Zhang, Yanzhao Zhang, Wen Xie, Mingxin Li, Ziqi Dai, Dingkun Long, Pengjun Xie, Meishan Zhang, Wenjie Li, and Min Zhang.
\newblock Gme: Improving universal multimodal retrieval by multimodal llms, 2025.
\newblock URL \url{https://arxiv.org/abs/2412.16855}.

\bibitem[Zhou et~al.(2017)Zhou, Zhao, Puig, Fidler, Barriuso, and Torralba]{Zhou_2017_CVPR}
Bolei Zhou, Hang Zhao, Xavier Puig, Sanja Fidler, Adela Barriuso, and Antonio Torralba.
\newblock Scene parsing through ade20k dataset.
\newblock In \emph{Proceedings of the IEEE Conference on Computer Vision and Pattern Recognition (CVPR)}, July 2017.

\bibitem[Zhou et~al.(2024)Zhou, Liu, Liu, Xiao, Wang, Zhao, Zhang, Lian, and Xiong]{megapairs}
Junjie Zhou, Zheng Liu, Ze~Liu, Shitao Xiao, Yueze Wang, Bo~Zhao, Chen~Jason Zhang, Defu Lian, and Yongping Xiong.
\newblock Megapairs: Massive data synthesis for universal multimodal retrieval, 2024.
\newblock URL \url{https://arxiv.org/abs/2412.14475}.

\bibitem[Zhu et~al.(2016)Zhu, Groth, Bernstein, and Fei-Fei]{visual7w}
Yuke Zhu, Oliver Groth, Michael Bernstein, and Li~Fei-Fei.
\newblock Visual7w: Grounded question answering in images, 2016.
\newblock URL \url{https://arxiv.org/abs/1511.03416}.

\end{thebibliography}
\bibliographystyle{tmlr}

\appendix

\section{Instruction Fine-tuning Prompt Design}
\label{apx:prompt}

We use the following prompt template when using GPT-4o to generate instructions:

\texttt{"<Image> Given this image, provide a user prompt where the following caption would be a reasonable answer: \{Caption\}. Only return the prompt."}

Where \texttt{\{Caption\}} is the caption for the bounding box from Visual Genome.
We use a temperature of 0 (deterministic) when generating instructions.
We choose this template to collect a diverse set of instructions that a user would plausibly ask.

\section{\texttt{CtrlBench} Prompt Design}
\label{apx:cbenchprompt}
\revised{We use the following template in Gemini 2.5 Flash for each of the 1000 images:}

\lstset{
  basicstyle=\ttfamily\small,
  breaklines=true,
  breakatwhitespace=true,
  frame=single,
  backgroundcolor=\color{gray!10},
  columns=fullflexible,
  keepspaces=true
}

\begin{lstlisting}
"<Image> Your task is to identify and describe 5 distinct objects or elements in this image. For each object provide a short sentence description of it. Please be specific in your description. Then, for each description, provide a user prompt where that description caption would be a reasonable answer. Output your answer as JSON in the following template: {"desc_1": <DESCRIPTION 1>, "prompt_1": <PROMPT 1>, "desc_2": <DESCRIPTION 2>, "prompt_2": <PROMPT 2>, "desc_3": <DESCRIPTION 3>, "prompt_3": <PROMPT 3>, "desc_4": <DESCRIPTION 4>, "prompt_4": <PROMPT 4>, "desc_5": <DESCRIPTION 5>, "prompt_5": <PROMPT 5>}
\end{lstlisting}

\revised{Save on API costs, we generate all the candidates and queries for each in a single prompt.}

\section{Architecture Ablation}
\label{apx:arch}

\begin{figure}[H]
    \centering

    \hfill
    \begin{minipage}[t]{0.40\textwidth}
    \includegraphics[width=\textwidth]{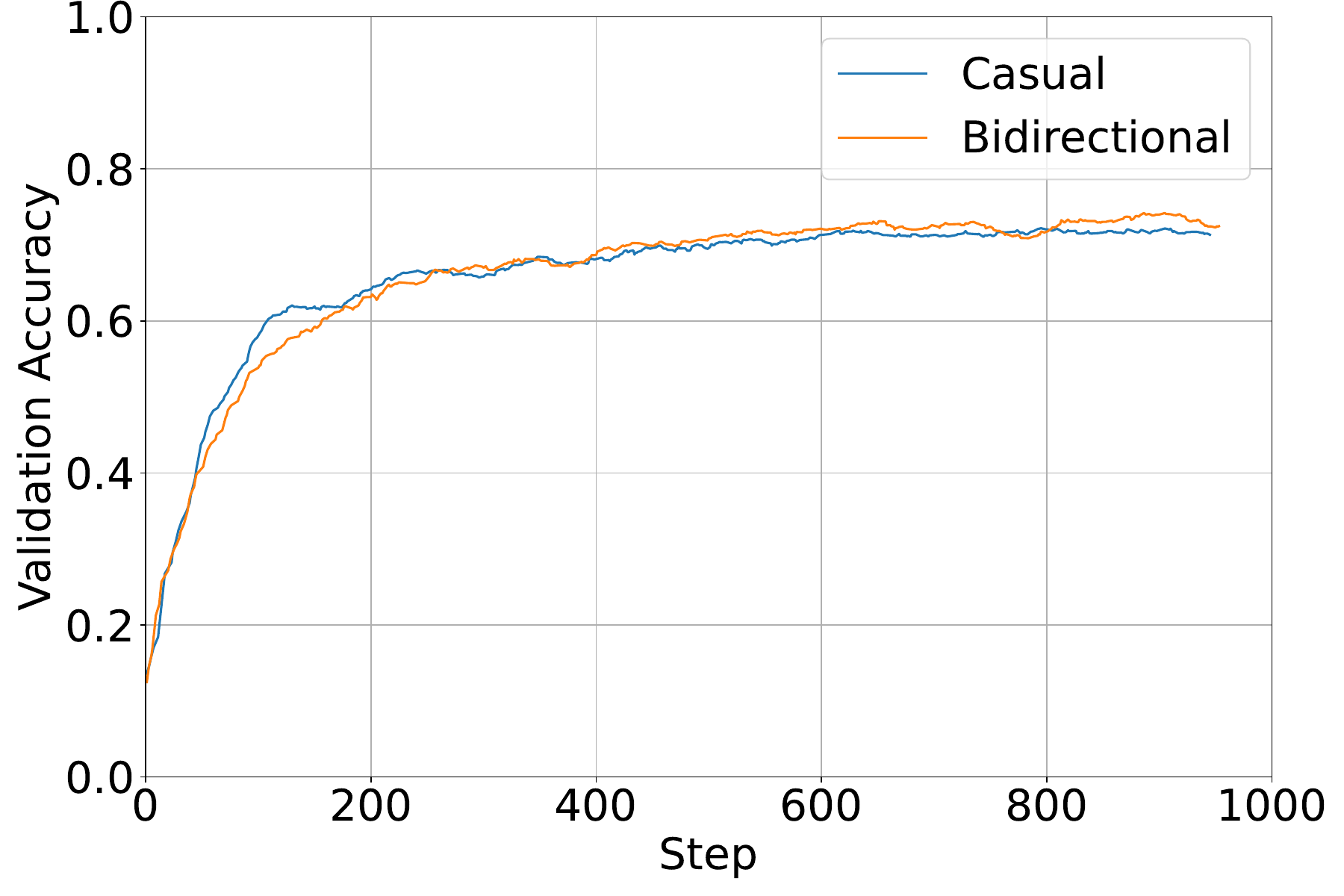}
    \end{minipage}
    \hfill
    \begin{minipage}[t]{0.40\textwidth}
        \centering
        \includegraphics[width=\textwidth]{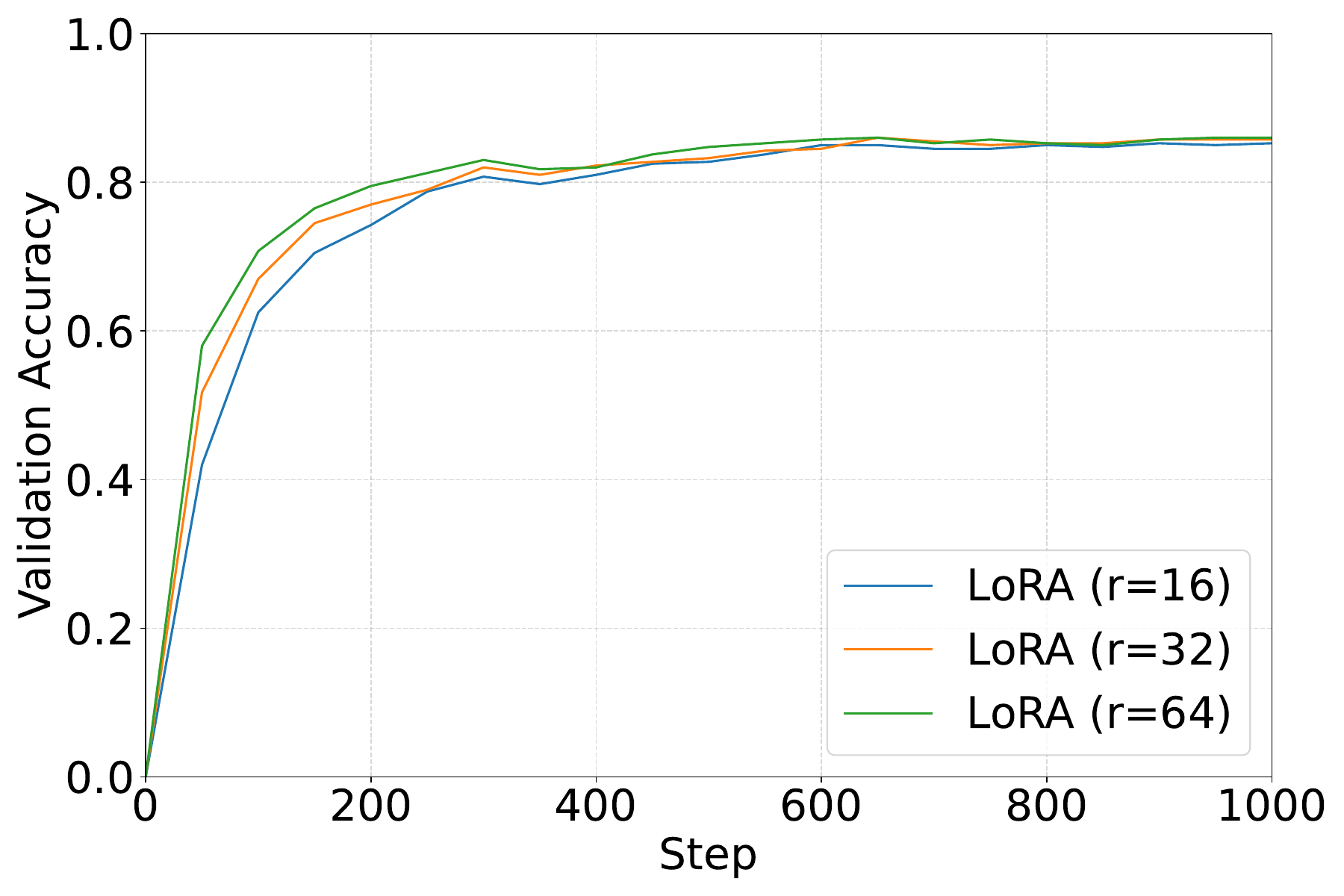}
    
    \end{minipage}
    \hfill
    \caption{\textbf{(Left)} Causal vs Bidirectional attention. \textbf{(Right)} LoRA adapter rank ablation.}
    \label{fig:arch}
\end{figure}

LLM2Vec~\citep{llm2vec} introduced several architecture modifications for adapting LLMs into text embedding models.
In the multimodal setting, we find that using casual vs. bidirectional attention is marginal for improving model accuracy~(\cref{fig:arch}).
On average, bidirectional attention \emph{slightly} outperforms casual attention.
Ablating over adapter rank, we find that different rank adapters tend to converge to the same accuracy, with higher rank adapters performing a fraction of a percentage better.
Overall, better data and scaling~(\cref{apx:scale}) present a much more promising direction for improving embedding quality.

\section{Scaling Training}
\label{apx:scale}

\begin{figure}
    \vspace{-0.5cm}
    \centering
    \includegraphics[width=0.45\linewidth]{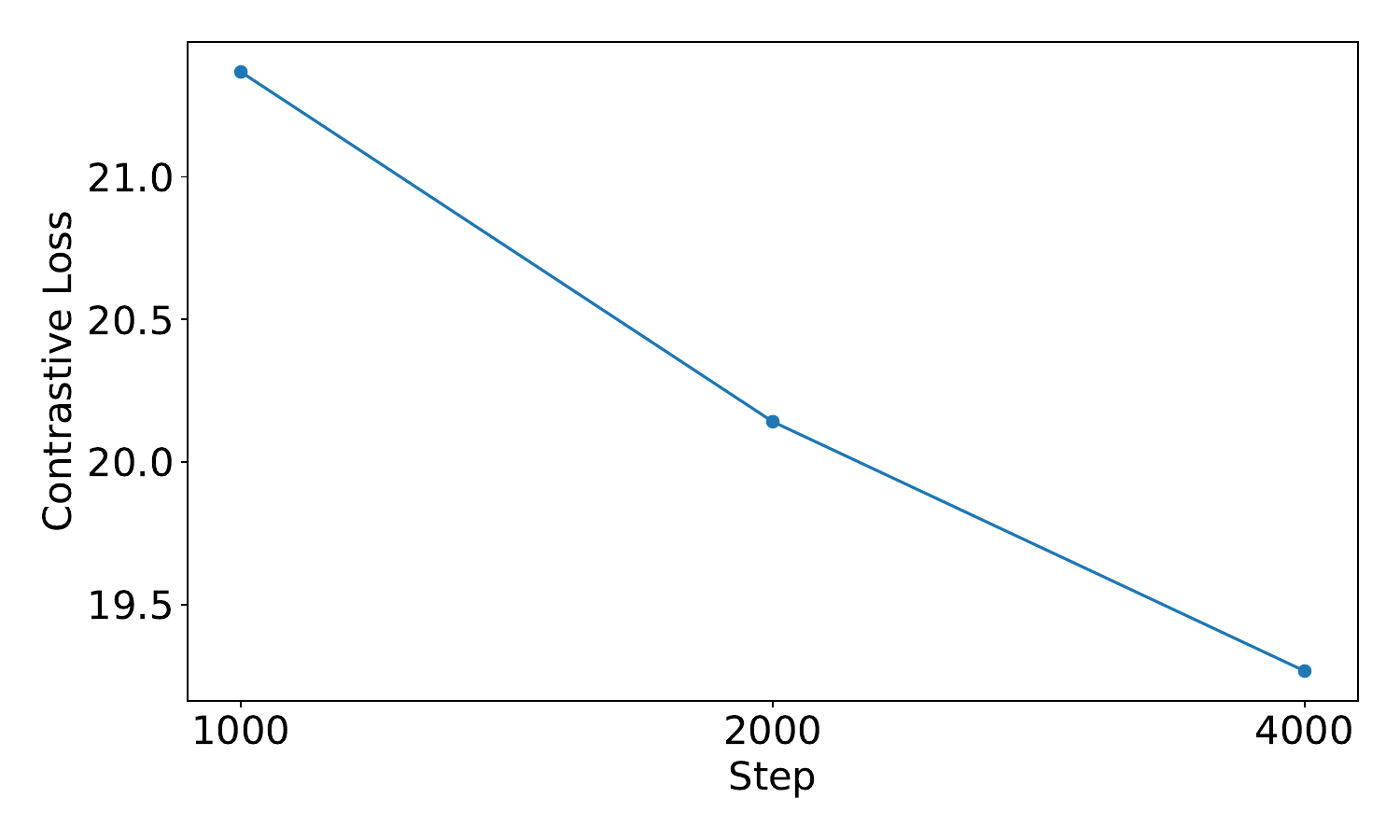}
    \caption{More samples seen vs. loss.}
    \label{fig:scaling_effects}
\end{figure}

We find that batch size and number of samples seen during training are both important factors.
By increasing our batch size by $4$x, from $128$ queries and $1024$ candidates to our full pretraining run size ($512$ queries and $4096$ candidates), our validation accuracy on an $800$ sample validation batch using in-batch negatives increases from $92.5\%$ to $95.0\%$.
We control for total samples seen by training the smaller batch size model for $4$x more steps.
Interestingly, this effect is even more pronounced on OOD validation data. For example, on a similar batch of MSCOCO data, accuracy increases from $73.8\%$ to $84.0\%$.
Therefore, techniques for scaling batch size under limited VRAM, such as GradCache~\cite{gradcache}, are a promising direction to improve our pretrained model.
Furthermore, we find that more steps (more samples seen) is also a straightforward way to increase the pretrained model's performance.
As shown by \cref{fig:scaling_effects}, doubling step count steadily decreases loss during our pretraining run.

\section{Comparison to VLM2Vec}
\label{apx:eval}
With the MMEB benchmark, \citet{vlm2vec} also include a mutltimodal embedding model: VLM2Vec.
We exclude VLM2Vec from \cref{tab:retrieval_results,tab:classification_results,tab:vqa_results} as it has been trained on MSCOCO and several tasks from the MMEB benchmark.
We provide the comparison of the two models here while specifying which tasks are out of distribution (OOD) for VLM2Vec.
All tasks are OOD for our model.

\begin{table}[H]
\centering
\begin{tabular}{l|cc}
& \textbf{VLM2Vec} & \textbf{ABC (ours)} \\
\midrule
\rowcolor{LightGray} ImageNet-1K~\citep{imagenet}    & 65.6  & \textbf{71.2} \\
\rowcolor{LightGray} HatefulMemes~\cite{hateful-memes}& \textbf{67.1} & 52.1 \\
\rowcolor{LightGray} VOC2007~\cite{voc-2007}         & \textbf{88.6}  & 81.4 \\
\rowcolor{LightGray} SUN397~\cite{sun397}            & \textbf{72.7}  & 71.8 \\
Place365~\cite{place365}        & \textbf{42.6}  & 40.7 \\
ImageNet-A~\citep{imagenet-a}   & 19.3 & \textbf{49.4} \\
ImageNet-R~\citep{imagenet-r}   & 70.2  & \textbf{86.8} \\
ObjectNet~\cite{objectnet}      & 29.5  & \textbf{67.7} \\
Country-211~\cite{clip}         & 13.0  & \textbf{18.5} \\
\midrule
\emph{Average}                  & 52.1  & \textbf{60.0} \\
\emph{Average OOD}              & 34.9  & \textbf{52.6} \\
\bottomrule
\end{tabular}
\caption{{Classification results} for {VLM2Vec} and \textbf{ABC (ours)}. Gray background indicates that {VLM2Vec} has been trained on this task.}
\label{tab:vlm2vec}
\end{table}

\cref{tab:vlm2vec} compares the performance of our model to VLM2Vec~\cite{vlm2vec}. Gray rows indicate tasks that {VLM2Vec} is trained on. On average, VLM2Vec performs better on tasks that are in its distribution, while our model outperforms on tasks where both models are OOD. Interestingly, our model performs better on ImageNet-1K even though it is not trained on it.

\section{Results on VLLMs without Contrastive Training}
\revised{Can dense embedding be extracted from VLLM's without additional contrastive training?
We experiment with two common methods for extracting dense embeddings from last-layer hiddens: \textbf{(1)} Using the last token in the sequence~\citep{vlm2vec}. \textbf{(2)} Averaging over the sequence dimension~\citep{llm2vec}.}

\begin{table}[h]
\centering
\resizebox{0.5\columnwidth}{!}{%
\begin{tabular}{llccc}
\toprule
\textbf{Pooling Method} & \textbf{Direction} & \textbf{Top-1} & \textbf{Top-5} & \textbf{Top-10} \\
\midrule
Mean Pooling & i2t & 6.30\% & 16.18\% & 23.70\% \\
             & t2i & 3.41\% & 9.25\%  & 13.39\% \\
Last Token   & i2t & 0.06\% & 0.26\%  & 0.36\% \\
             & t2i & 0.08\% & 0.30\%  & 0.73\% \\
\bottomrule
\end{tabular}%
}
\caption{Comparison of pooling methods across retrieval directions and on MSCOCO.}
\label{tab:notrain_pooling_comparison}
\end{table}

\revised{On MSCOCO retrieval we find that mean pooling significantly outperforms using the last token hidden (\cref{tab:notrain_pooling_comparison}).
However, neither method performs very strong.
Both methods are significantly worse than a zero-shot CLIP model on MSCOCO (\cref{tab:retrieval_results}).}

\section{Inference Settings for CLIP}
\label{apx:infernece}

\begin{table}[H]
\centering
\resizebox{0.38\columnwidth}{!}{%
\begin{tabular}{lcc}
\toprule
\textbf{Resolution} & \textbf{Image $\rightarrow$ Text} & \textbf{Text $\rightarrow$ Image} \\
\midrule
224x224 & 64.94 & 48.02 \\
336x336 & 63.28 & 47.42 \\
448x448 & 61.98 & 45.97 \\
\bottomrule
\end{tabular}%
}
\caption{OpenClip-ViT-g-14 accuracy (top-1) on MSCOCO at different image resolutions. Unlike \textbf{ABC}, accuracy degrades when scaling token budgets beyond OpenClip-ViT-g-14's native $224\times224$ resolution.}
\label{tab:clip_resolution}
\end{table}
\vspace{-0.25cm}

We find that CLIP models perform best at their pretraining resolution. As shown in \Cref{tab:clip_resolution} accuracy consistently degrades when scaling the token counts of CLIP models beyond their native resolution. It would be unfair to evaluate models at a setting that is both more compute intensive and less performant. Therefore, we evaluate CLIP models at their native resolution and token count across all experiments.

\newpage
\section{\texttt{CtrlBench Examples}}
\label{apx:ctrlbench}

\begin{figure}[H]
    \centering
    \captionsetup[subfigure]{font=footnotesize}
    
    \begin{subfigure}[t]{0.48\textwidth}
        \centering
        \includegraphics[width=\textwidth,height=0.5\textheight,keepaspectratio]{ctrlbench/img3.jpg}
        \begin{footnotesize}
            \begin{itemize}[label={}]
                \item \textbf{Q}: What is happening in the top left corner of the image? \\ \textbf{A}: Tennis ball in the air
                \item \textbf{Q}: What is the player about to hit with his racket? \\ \textbf{A}: A round tennis ball
                \item \textbf{Q}: What color are the shorts the tennis player is wearing? \\ \textbf{A}: black and yellow shorts
                \item \textbf{Q}: What action is taking place on the tennis court in the image? \\ \textbf{A}: A man serving a tennis ball.
                \item \textbf{Q}: What's prominently dividing the ground area in the image? \\ \textbf{A}: White line on a court
            \end{itemize}
        \end{footnotesize}
        \label{fig:img3}
    \end{subfigure}
    \hfill
    \begin{subfigure}[t]{0.48\textwidth}
        \centering
        \includegraphics[width=\textwidth,height=0.5\textheight,keepaspectratio]{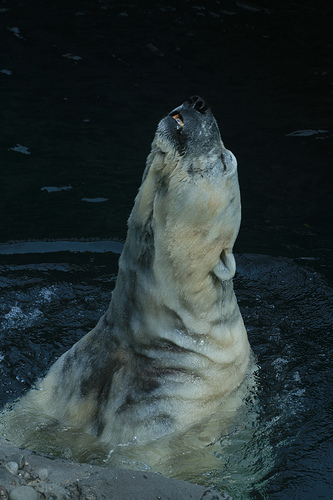}
        \begin{footnotesize}
            \begin{itemize}[label={}]
                \item \textbf{Q}: What part of the animal is prominently displayed in the image? \\ \textbf{A}: the neck of a polar bear
                \item \textbf{Q}: What colors can you see on the bear in the image? \\ \textbf{A}: black, brown and white bear neck
                \item \textbf{Q}: What stands out to you about the bear's appearance in this image? \\ \textbf{A}: the bear has such tiny ears for such a huge animal
                \item \textbf{Q}: What is the bear doing with its eyes in this picture? \\ \textbf{A}: the bear has his eyes closed
                \item \textbf{Q}: What's the polar bear swimming in? \\ \textbf{A}: a body of water
            \end{itemize}
        \end{footnotesize}
        \label{fig:img4}
    \end{subfigure}
    \label{fig:vqa_part2}
\end{figure}

\end{document}